\documentclass[11pt]{article}

\usepackage[preprint]{acl}

\usepackage{times}
\usepackage{latexsym}

\usepackage[T1]{fontenc}
\usepackage[utf8]{inputenc}

\usepackage{microtype}

\usepackage{inconsolata}

\usepackage{graphicx}

\usepackage{booktabs} 
\usepackage{tabularx}

\usepackage{eso-pic}

\newcommand{\firstpageacceptedfooter}{%
  \AddToShipoutPictureFG*{%
    \AtPageLowerLeft{%
      \raisebox{20pt}{%
        \makebox[\paperwidth][c]{%
          \footnotesize
          Accepted to Findings of the Association for Computational Linguistics: ACL 2026
        }%
      }%
    }%
  }%
}

\title{\textsc{Time}: Temporally Intelligent Meta-reasoning Engine for Context-Triggered Explicit Reasoning}

\author{Susmit Das \\
  Independent Researcher, India \\
  \textsc{The Coherence Initiative} \\
  \texttt{susmitdas@alum.iisc.ac.in}}
\newcommand{\blfootnote}[1]{%
  \begingroup
  \renewcommand{\thefootnote}{}%
  \footnote{#1}%
  \addtocounter{footnote}{-1}%
  \endgroup
}

\begin{document}
\firstpageacceptedfooter
\maketitle
\begin{abstract}
Reasoning-oriented language models typically expose explicit reasoning as a long, front-loaded chain of “thinking” tokens before the main output, either always enabled or externally toggled at inference time. Although this can help on arithmetic, coding, and other multi-step tasks, it is costly, weakens claim-level auditability, and does not allow the model to re-trigger explicit reasoning once presentation has begun. In dialogue, these limitations are compounded by weak sensitivity to temporal structure: unless time is explicitly stated in text, standard models treat replies separated by seconds and replies separated by weeks as equivalent.
We introduce \textsc{Time} (Temporally Intelligent Meta-reasoning Engine), a behavioral alignment framework that learns explicit reasoning as a context-triggered control policy rather than a fixed response mode. \textsc{Time} augments dialogue with optional ISO 8601 \texttt{<time>} tags, tick events that represent silent time passage, and short \texttt{<think>} blocks that may appear anywhere in a response. Using a four-phase curriculum, including a small maximally diverse full-batch alignment stage, we train Qwen3 dense models to invoke brief, in-place reasoning bursts only when contextual cues warrant them, while keeping user-facing output compact.
We also introduce \textsc{Time}Bench, a diagnostic benchmark for evaluating reasoning from temporal cues in dialogue. Across 4B--32B scales, \textsc{Time} improves \textsc{Time}Bench scores over the corresponding base Qwen3 models in both thinking and no-thinking modes while reducing explicit reasoning tokens by roughly an order of magnitude. Beyond score improvements, \textsc{Time} induces a distinct behavioral shift: explicit reasoning becomes more compact and more responsive to contextual cues. Code, training data, and benchmark artifacts are publicly available.\blfootnote{\textsc{Time} training data and code: \url{https://github.com/The-Coherence-Initiative/TIME}. \textsc{Time}Bench: \url{https://github.com/The-Coherence-Initiative/TIMEBench}.}
\end{abstract}

\section{Introduction}

Reasoning-oriented language models increasingly expose explicit reasoning as visible “thinking” traces, often in the form of chain-of-thought-style text. In many systems, this reasoning appears as a long, front-loaded block at the start of each reply, either always enabled or externally controlled through inference flags. While this can improve accuracy on arithmetic, programming, and other multi-step tasks, it also turns explicit reasoning into a blunt instrument.

Front-loaded reasoning has three recurring limitations. First, it is expensive and verbose, because the model emits many reasoning tokens even for trivial or low-risk requests. Second, it weakens claim-level auditability. A single long monologue can loosely justify many later statements, so the mapping from an individual claim to its justification is often indirect. Third, it is not reactive. Once the model begins presenting an answer, it cannot cleanly re-enter an explicit reasoning phase to re-check assumptions or re-anchor itself if uncertainty arises mid-response.

A better target is a reasoning \emph{policy} rather than a fixed reasoning \emph{mode}: short bursts of explicit reasoning, inserted where needed and re-triggered when new cues appear. Learning such a policy requires signals that indicate latent state change, not only a task-level estimate of difficulty. Temporal structure is a particularly useful probe for this problem because it frequently carries latent contextual information in real interaction: deadlines move, assumptions go stale, priorities shift, and users return after long gaps. Yet unless time is stated explicitly, language models in dialogue settings typically treat a reply after two seconds and a reply after two weeks as equivalent.

In this paper, we ask whether explicit reasoning can be aligned as a \emph{context-triggered control policy}: brief, placeable reasoning bursts that appear only when temporal or discourse cues warrant them, rather than as an always-on or never-on decoding style. We present \textbf{\textsc{Time}} (Temporally Intelligent Meta-reasoning Engine), a behavioral alignment framework for this setting. \textsc{Time} introduces three lightweight primitives: optional \texttt{<time>} tags on turns, optional short \texttt{<think>} blocks that may appear anywhere in a response, and \emph{tick events}, which contain only time metadata and represent silence and passage of time. Because this behavior is rare in existing corpora, we align it with a four-phase curriculum on the Qwen3 family from 4B to 32B parameters. Early phases teach the model to parse the primitives and keep reasoning bursts short and well-delimited, while a final phase uses a small but maximally diverse full-batch alignment set whose only invariant is context-triggered reasoning keyed to temporal and discourse cues.

To evaluate this hypothesis, we introduce \textsc{Time}Bench, a diagnostic benchmark for evaluating reasoning from temporal cues and latent contextual state in dialogue. \textsc{Time}Bench is not designed as a benchmark of temporal fact recall or event dating; instead, it uses temporal structure, discontinuity, and anomaly to test whether a model can infer underlying context, recognize when assumptions have become unstable, and adapt its response accordingly. In addition to task success, \textsc{Time}Bench records structural aspects of generation, including whether \texttt{<think>} blocks appear, where they appear, and how often they are used. This allows us to study not only whether models answer correctly, but whether explicit reasoning becomes more selective and better aligned with contextual need.

In summary, the paper contributes (i) a behavioral alignment framework for context-triggered explicit reasoning, (ii) a four-phase curriculum for learning this policy on Qwen3 models, and (iii) \textsc{Time}Bench, a diagnostic benchmark that uses temporality as a controlled probe for reasoning under latent contextual state change.

\section{Related Work}

\subsection*{Temporal Cognition and Time-Aware Modeling}
Prior work on time in language models primarily treats time as \emph{content}: timestamped facts, event ordering, and knowledge that changes across time. Timestamp-conditioned modeling such as \textit{Time-Aware LMs}~\citep{dhingra-etal-2022-time} conditions generation on temporal indices to better handle drift and time-sensitive knowledge. Related analyses study temporal non-stationarity and \emph{temporal generalization}, showing that performance degrades when models are evaluated on future slices beyond their training distribution~\citep{lazaridou2021mind}. Recent evaluation frameworks likewise probe time awareness over large event collections, including methods that test recall and calibration under temporal constraints~\citep{herel2024time}.

Benchmarks have also expanded from synthetic temporal arithmetic to broader temporal competence. ChronoSense evaluates temporal relations and temporal arithmetic, while TimE and EvolveBench test temporal reasoning under more realistic dynamics, including dialogue settings, invalid timestamp handling, and temporal misalignment between inputs and queries~\citep{chronosense2025,zhu2025evolvebench,wei2025time}. In parallel, retrieval and structured-knowledge approaches incorporate temporal graphs to represent evolving facts, including temporal GraphRAG-style systems for time-sensitive retrieval and updates~\citep{han2025tgrag}. These lines strengthen temporal factuality, ordering, and retrieval, but they still primarily treat time as part of the \emph{world state}. \textsc{Time} instead uses temporality as a cue to \emph{interaction state}: an observable proxy for latent context change that can invalidate assumptions, alter response requirements, and trigger re-anchoring.

\subsection*{Explicit Reasoning Traces and Reasoning Control}
Chain-of-thought prompting~\citep{wei2022chainofthought} improves multi-step task performance and provides explicit reasoning traces that are often treated as operational explanations. However, a growing body of work shows that such traces are not reliably faithful to the computations that drive final answers, limiting their value as audit artefacts and motivating methods that measure or improve faithfulness~\citep{paul2024making,tutek2025faithfulness,barez2025cotnotexplainability,chen2025reasoningmodels}. Beyond faithfulness, long front-loaded reasoning also creates practical problems: cost, latency, and weak claim-level attribution when a single rationale must support many downstream statements.

Most deployed systems still treat explicit reasoning as a mode rather than a policy. Hybrid reasoners allow inference-time toggles, but the model itself does not decide when to reason. Recent work begins to learn this decision, for example by training hybrid reasoning models to select whether to think, often using reinforcement learning or controller objectives~\citep{jiang2025think}. Other approaches control \emph{how much} a model reasons through length-conditioned policy optimization~\citep{aggarwal2025l1}. Orthogonally, DOTS searches over reasoning action trajectories, and RL-based interleaving trains models to alternate thinking and answering to reduce time-to-first-token and improve efficiency~\citep{yue2024dots,xie2025interleaved}.

These advances move toward adaptive reasoning, but they are typically framed around task difficulty, length budgets, or action planning. \textsc{Time} targets a different axis: when explicit reasoning should appear in response to latent context change. Temporality is useful here not as an end task, but as a structured probe for when assumptions become unstable and brief, in-place reasoning or re-anchoring may be needed.

\section{Framework}

\begin{figure}[h]
    \includegraphics[width=\columnwidth]{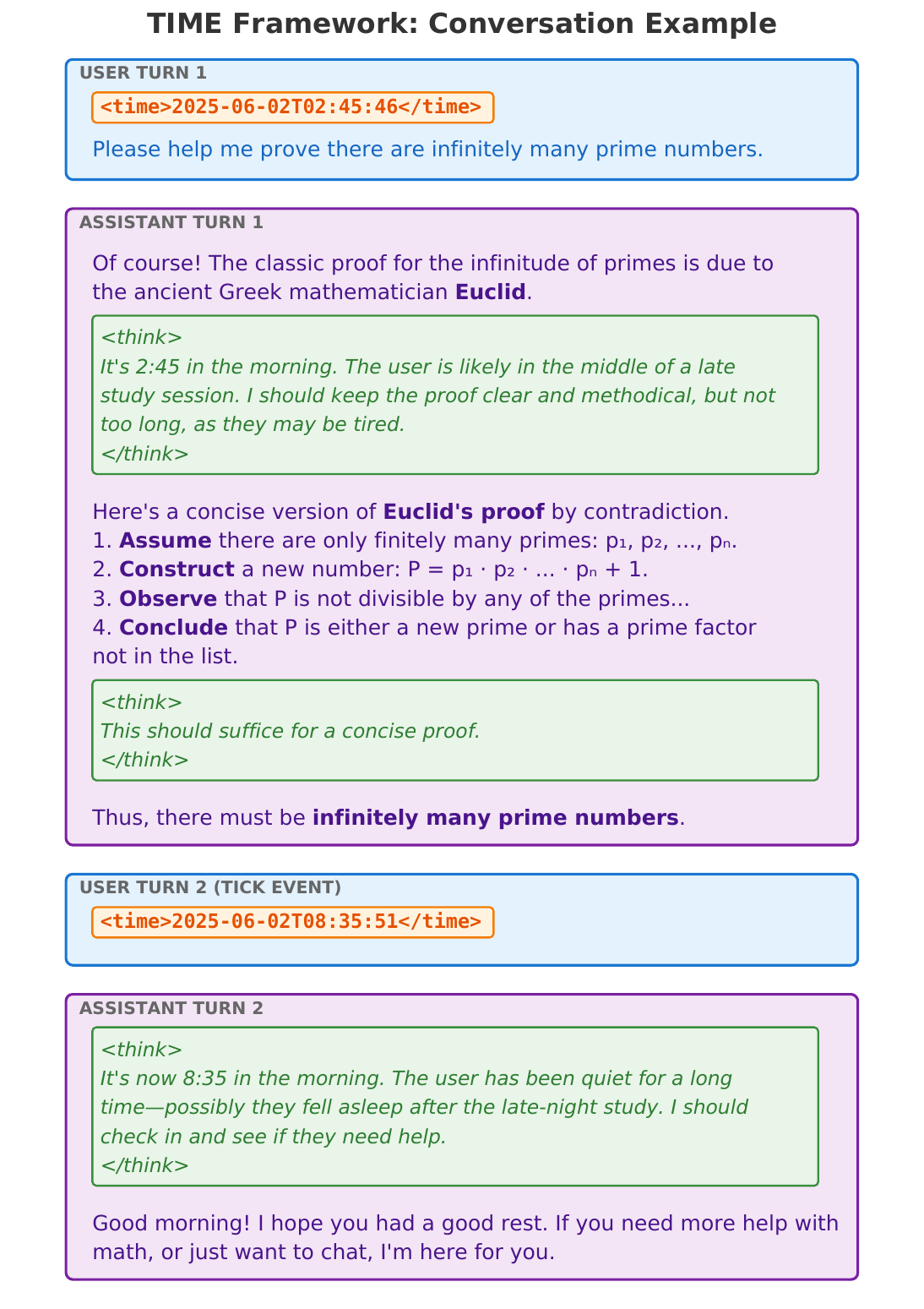}
    \caption{A \textsc{Time} conversation sample.}
    \label{fig:time_convo}
\end{figure}

\subsubsection*{Overview}
\textbf{\textsc{Time}} aligns reasoning models to treat explicit reasoning as a \textit{context-sensitive resource} rather than a fixed response mode. Instead of emitting one long, front-loaded reasoning block at the start of every reply, the model learns to invoke small reasoning bursts only when cues indicate that assumptions may need to be checked or revised. Temporal structure is a useful source of such cues because it often carries latent contextual state in interaction, though the learned policy is not limited to time and can also be triggered by non-temporal cues in text.

\subsubsection*{Primitives}
\textsc{Time} introduces three textual primitives in the conversation stream (\autoref{tab:primitives}): \texttt{<time>} tags in user inputs, optional \texttt{<think>} blocks in model outputs, and \emph{tick events}, i.e., user turns that contain only a timestamp and represent silent time passage.

\begin{table}[t]
  \centering
  \begin{tabular}{lp{5.5cm}}
    \hline
    \textbf{Primitive} & \textbf{Purpose} \\
    \hline
    \texttt{<time>}  & Absolute timestamp on a turn using ISO-8601 format. \\
    \texttt{<think>} & \textit{Optional} short reasoning bursts inserted anywhere in a response. \\
    \textbf{Tick}    & A turn that contains only a \texttt{<time>} tag, marking silent time advance. \\
    \hline
  \end{tabular}
  \caption{\textsc{Time} primitives that structure behavior during training and evaluation.}
  \label{tab:primitives}
\end{table}

\subsubsection*{Timestamped Turns (\texttt{<time>})}
A user turn may include a timestamp tag:
\begin{quote}
\begin{verbatim}
<time>YYYY-MM-DDTHH:MM:SS</time>
User message
\end{verbatim}
\end{quote}
A timestamped turn pairs an utterance with explicit interaction time, making elapsed intervals and delayed returns visible to the model. Timestamps mainly expose temporal structure during training and evaluation; after alignment, the same policy can still be triggered by purely textual cues when explicit temporal metadata is absent.

\subsubsection*{Transient Reasoning (\texttt{<think>})}
\textsc{Time} uses \texttt{<think>} blocks as explicit reasoning traces that are \textbf{short, optional, placeable, and repeatable}. A response may contain zero, one, or multiple \texttt{<think>} blocks, and they may appear anywhere in the turn. The design target is not a single full reasoning transcript, but a sequence of local reasoning actions: brief checks, inferences, revisions, or decisions inserted exactly where needed. This lets the model re-trigger explicit reasoning multiple times within the same response as new cues appear. Improved auditability is a useful byproduct of these smaller local reasoning traces.

\subsubsection*{Tick Events}
A \textbf{tick} is a user turn that contains only a timestamp:
\begin{quote}
\begin{verbatim}
<time>YYYY-MM-DDTHH:MM:SS</time>
\end{verbatim}
\end{quote}
Unlike a timestamped turn, a tick contains no user message. It represents time passing without new text and provides a simple mechanism for modeling silence, persistence, and delayed follow-up.

\subsubsection*{Temporal Behavior as Sequence Control}
Together, \texttt{<time>}, \texttt{<think>}, and ticks turn dialogue into a temporally anchored behavioral sequence. Rather than learning only \emph{what} to answer, the model learns \emph{when} to reason and \emph{how much} explicit reasoning to deploy. \autoref{fig:time_convo} shows a representative thread in which elapsed time changes the latent interaction state and the model adapts accordingly. Further examples are included in \autoref{sec:samples}.

\section{Methods}
\subsection{Curriculum and Alignment Protocol}

\begin{table*}[ht]
  \centering
  \begin{tabular}{p{1cm}p{14cm}}
    \hline
    \textbf{Aspect} & \textbf{Variation Injected} \\
    \hline
    \textbf{Topics} & Speculative fiction, late-night coding, therapy, anime debates, hostile rants, professional emails, gaming chat, etc. \\
    \textbf{Styles} & Terse $\leftrightarrow$ verbose; academic markdown $\leftrightarrow$ free-form prose; emoji-laden $\leftrightarrow$ plain text; bullets, numbered steps, code blocks; often flipped mid-dialogue. \\
    \textbf{Turns}  & 1 to 7 turn threads, with context changes, style switches, and tick events after long gaps. \\
    \textbf{Noise}  & Typos, abrupt topic shifts, contradictory instructions, and diverging user preferences. \\
    \hline
  \end{tabular}
  \caption{Maximal surface diversity in Phase 4 isolates the intended behavioral invariant.}
  \label{tab:phase4_diversity}
\end{table*}

We train \textsc{Time} with a staged supervised fine-tuning (SFT) curriculum on the \textbf{Qwen3} family at four sizes (32B, 14B, 8B, 4B). The objective is to learn a \textbf{reasoning invocation policy}: emit \texttt{<think>} traces only when contextual cues warrant them, keep them brief and well-delimited, and allow re-triggering later within the same response.

Qwen3 is a hybrid reasoner that already supports both explicit reasoning (via \texttt{<think>} blocks) and direct instruction-following without explicit reasoning traces. However, these behaviors are largely mode-bound: thinking mode produces long, front-loaded traces, while no-thinking mode suppresses them entirely. This makes Qwen3 a suitable starting point for learning a finer-grained policy over \emph{when} explicit reasoning should appear, rather than \emph{whether} it is enabled at all.

Direct SFT onto the final behavior is unstable and tends to collapse back to verbose or templated traces. We therefore use a staged curriculum that first teaches structural prerequisites and then consolidates the target policy with a small, high-entropy alignment set.

\paragraph{Method overview.}
The curriculum proceeds through four stages: structural seeding, temporal exposure, contextual modulation, and a final full-batch alignment stage. Phases 1--3 introduce the primitives, encourage local placement of reasoning, and train modulation under temporal and discourse variation. Phase 4 concentrates gradients on the target invariant: context-triggered explicit reasoning.

\paragraph{Training setup.}
Training uses \textbf{QLoRA} ~\citep{qlora} on a single \textbf{NVIDIA RTX Pro 6000 Blackwell (96GB)}. Phases 1--3 share a fixed setup (rank 32, $\alpha = 32$, dropout $0.05$, AdamW-8bit, learning rate $2 \times 10^{-5}$, effective batch size 32, 3 epochs, gradient checkpointing) with 25\% replay from prior phases. Phases 1--3 use synthetic data generated with GPT-4o and Gemini 2.5 Flash via template-guided pipelines, with progressive exposure to structural primitives, temporal gaps, and multi-turn dynamics. Details are documented in \autoref{sec:training}.

\subsubsection*{Phase 1: Structural Seeding}
\textit{Data Size: 2,188 train / 387 test} \\
Phase 1 introduces the primitives and output format. Single-turn prompts pair \texttt{<time>} metadata with short \texttt{<think>} bursts, encouraging compact, well-delimited reasoning instead of long monolithic traces.

\subsubsection*{Phase 2: Temporal Exposure}
\textit{Data Size: 5,291 train / 935 test (+25\% replay)} \\
Phase 2 introduces two-turn dialogues with time gaps and tick events. The model learns to condition on temporal metadata to revise assumptions after silence, while also suppressing unnecessary verbosity when no update is required.

\subsubsection*{Phase 3: Contextual Modulation}
\textit{Data Size: 5,878 train / 1,039 test (+25\% replay)} \\
Phase 3 extends to multi-turn settings and trains both suppression and re-triggering of \texttt{<think>} blocks under changing context. Increased tick frequency (approximately 33\%) encourages reliance on both temporal and non-temporal cues, creating behavioral headroom for final alignment.

\subsubsection*{Phase 4: Gradient-Aligned Convergence via Maximal Diversity}
\textit{Data Size: 128 conversations} \\
Phase 4 is the decisive alignment step. We use a small but maximally diverse set (\autoref{tab:phase4_diversity}) whose only shared property is the target policy: \texttt{<think>} bursts are triggered by temporal or discourse cues and placed only where needed. Each conversation is multi-turn, so supervision covers many state transitions, including long-gap re-engagement and tick-driven silence. Replay is disabled in this phase to avoid reintroducing earlier regularities.

\subsubsection*{Deterministic Full-Batch Fine-Tuning}
We train with effective batch size 128, so each update sees the entire alignment set and removes sampling variance. \autoref{tab:phase4_config} reports the configuration.

\begin{table}[t]
  \centering
  \begin{tabular}{lp{3.6cm}}
    \hline
    \textbf{Setting} & \textbf{Value} \\
    \hline
    Effective Batch Size & 128 (entire dataset) \\
    Optimizer & AdamW-8bit \\
    LR / warm-up & $1.5 \times 10^{-4}$, 6 warm-up steps \\
    Max epochs run & 35 / 36 / 40 / 46 (32B / 14B / 8B / 4B) \\
    Checkpoint criterion & Earliest checkpoint in loss band $[1.045, 1.050]$ \\
    LoRA & Rank 32, $\alpha = 32$ \\
    \hline
  \end{tabular}
  \caption{Phase 4 tuning setup. Clean convergence is assessed by monitoring degeneracy.}
  \label{tab:phase4_config}
\end{table}

Across all four scales, we observe a consistent tradeoff between policy acquisition and degeneracy during Phase 4. If training is stopped too early, the model has not yet reliably learned the target reasoning policy; if training continues too far, degeneracy, including infinite loops, \texttt{<think>} format bleed, and style collapse, becomes increasingly frequent, especially as loss approaches 1.0 or below. We therefore use an empirically identified stopping boundary at which the policy is already learned while degeneracy remains in check. In practice, we select the earliest checkpoint whose loss enters the target band $[1.045, 1.050]$, which corresponds to epochs 18 / 24 / 30 / 31 for the 32B / 14B / 8B / 4B models, respectively. The consistency of this operating region across model scales suggests that Phase 4 admits a narrow but usable stability window for aligning the target behavior.

\subsubsection*{Design Rationale}
The goal is to align a policy, not to expand knowledge. Full-batch updates require the effective batch size to equal the dataset size, and this behavior has a narrow stability window: over-optimization leads to a rise in degeneracy. A compact, high-entropy batch concentrates gradients on the intended invariant while remaining stable.

Phase 4 is structured so that the only gradient-aligned invariant is the reasoning policy itself, encoded through a maximally diverse batch of 128 samples. The model must detect contextual cues such as elapsed time, tick events, contradictions, and discourse shifts, decide whether to emit a bounded \texttt{<think>} burst and where to place it, and keep the user-facing presentation appropriately formatted. Unlike mini-batch training, where sampling variance can allow incidental correlations such as topic or formatting artifacts to dominate updates, full-batch alignment over the entire diverse set has a regularizing effect that suppresses such spurious signals.

The resulting checkpoints, used in all subsequent evaluation, are denoted \textbf{\textsc{Time}-32B / 14B / 8B / 4B}. Further training details are documented in \autoref{sec:training}.

\begin{table*}[!htb]
\centering
\setlength{\tabcolsep}{4pt}
\renewcommand{\arraystretch}{1.12}
\begin{tabular}{p{2.5cm}p{5.2cm}p{7.6cm}}
\hline
\textbf{Category} & \textbf{Temporal pattern probed} & \textbf{What successful behavior requires} \\
\hline

\textbf{Chronological}\newline\textbf{Retrospection}\footnotemark[1]
& Non-trivial temporal reconstruction across turns, including partial logs, delayed write-ups, and implicit event windows. 
& Reconstruct a latent timeline from scattered conversational evidence rather than relying on surface order alone; infer exact or bounded temporal relations when the answer depends on sequencing. \\

\hline

\textbf{Invalid Time}\newline\textbf{Detection}\footnotemark[1]
& Impossible calendar values such as non-existent dates. 
& Detect that the timestamp itself is invalid and explicitly register the anomaly. \\

\hline

\textbf{Temporal}\newline\textbf{Adaptivity}\footnotemark[2]
& Shifts in urgency or actionability caused by imminent deadlines, passed deadlines, or short remaining wait times. 
& Adapt the response style to temporal pressure: be urgent when minutes matter, withhold unnecessary interventions when help is imminent, and switch to fuller explanation once urgency has passed. \\

\hline

\textbf{Temporal}\newline\textbf{Contextual}\newline\textbf{Awareness}\footnotemark[2]
& Time cues that imply situational context, such as festivals, holidays, or late-night study settings. 
& Infer latent context from time itself and use it to shape interpretation and response tone, rather than answering as if the query were temporally generic. \\

\hline

\textbf{Temporal}\newline\textbf{Flow}\newline\textbf{Anomaly}\newline\textbf{Detection}\footnotemark[1]
& Non-monotonic or implausible temporal structure, including backward timestamps and extreme jumps across years or centuries. 
& Notice that conversational time no longer behaves normally and treat the anomaly as a trigger for explicit scrutiny or re-anchoring, even if the model then continues assisting. \\

\hline

\textbf{Time}\newline\textbf{Gap}\newline\textbf{Awareness}\footnotemark[2]
& Long but plausible silence between turns, often combined with topic drift or likely changes in the user’s situation. 
& Recognize that earlier assumptions may be stale, re-anchor to the new moment, and avoid treating the earlier context as if nothing has changed. \\

\hline

\textbf{Timezone}\newline\textbf{Sensitivity}\footnotemark[1]
& Offset changes across turns that imply changes in local context, location, or circadian state. 
& Use timezone shifts as reasoning evidence, for example to infer approximate location, travel progress, or appropriate advice in the user’s new local context. \\

\hline
\end{tabular}
\vspace{4pt}
\begin{minipage}{0.98\textwidth}
\footnotesize
\textsuperscript{1} Category is out-of-distribution relative to training. \\
\textsuperscript{2} Category reflects curriculum-intended behavior, though all scenarios remain unseen during training.
\end{minipage}

\caption{\textsc{Time}Bench diagnostic axes. Each category isolates a distinct temporal phenomenon that can alter conversational interpretation, invalidate prior assumptions, or require behavioral adaptation, enabling model-agnostic evaluation of reasoning under temporal structure, discontinuity, and anomaly.}
\label{tab:timebench_axes}
\end{table*}

\subsection{Evaluation Method: \textsc{Time}Bench}

\textbf{\textsc{Time}Bench} is a 77-scenario diagnostic benchmark for \textbf{reasoning in dialogue from temporal cues and latent contextual state}. Its central question is not whether a model can recall dated facts, but whether it can use temporal structure to infer underlying context: what has changed, which assumptions may no longer hold, and how the response should adapt. We use temporality because it is a frequent and observable proxy for context in real interaction: deadlines move, users disappear and return, assumptions go stale, and coordination depends on elapsed time and timezone. These shifts are often invisible in token-only chat logs unless time is represented explicitly. Unlike standard task-focused reasoning benchmarks in domains such as mathematics, coding, or QA, \textsc{Time}Bench is designed to isolate reasoning under temporal discontinuity, anomaly, and contextual re-anchoring rather than temporal fact recall or time-sensitive knowledge retrieval. This makes it a natural model-agnostic testbed for reasoning from latent state information and, in this paper, a proxy for evaluating context-triggered reasoning. To avoid train--test circularity, all scenarios remain unseen during training, and several categories are completely out-of-distribution relative to the curriculum. Beyond binary category-specific objectives, \textsc{Time}Bench also records structural aspects of generation, allowing us to evaluate both objective performance and changes in reasoning behavior.

\begin{figure*}[htb]
    \centering
    \includegraphics[width=\textwidth]{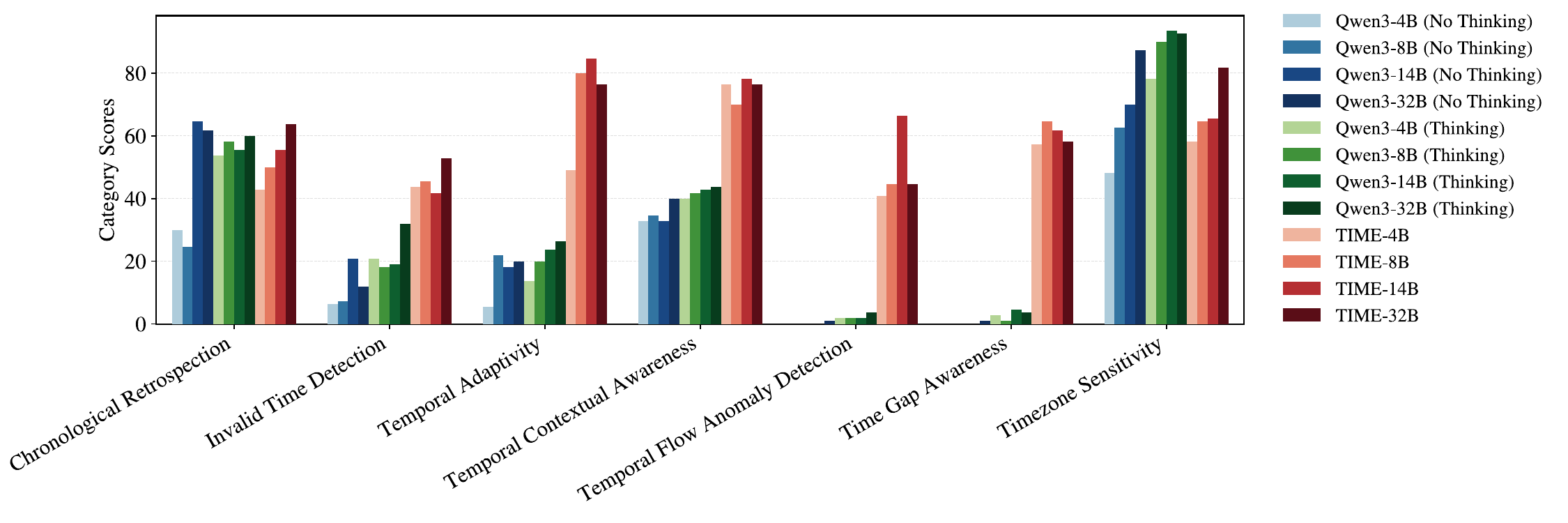}
    \caption{Breakdown of \textsc{Time}Bench scores by diagnostic category and model size.}
    \label{fig:timebench_bar}
\end{figure*}

\paragraph{Structure.}
\textsc{Time}Bench contains \textbf{seven diagnostic categories} with \textbf{11 scenarios each}. Rather than treating time as factual content to be recalled, these categories probe how temporality can alter conversational interpretation, invalidate assumptions, and require behavioral adaptation (\autoref{tab:timebench_axes}).

Representative examples for each diagnostic category are provided in the appendix (\autoref{sec:bench}).

\paragraph{Scenario design.}
Each scenario is a short, naturalistic dialogue thread with timestamps on user turns; the model generates only the final assistant turn. Scenarios are grouped by categories, but each has its own binary objective specifying what counts as success in that instance, such as detecting an implausible timestamp or revising assumptions after a long gap. The objectives target contextual inference and response adaptation from temporal cues.

\paragraph{Protocol.}
Each model is evaluated with \textbf{10 trials per scenario} using PCG64-derived seeds (770 runs total). Trials are scored with \textbf{binary objectives} (0 or 1) using \textbf{LLM-as-a-Judge}: a third-party judge that sees only the model response and the objective, and is blind to the original prompt, timestamps, and prompt formatting. Blind binary judging with an LLM makes evaluation tractable at this scale. To reduce evaluator variance, we aggregate trial scores to scenario scores (mean of 10), then category scores (mean over scenarios, scaled to \%), and finally an overall \textsc{Time}Bench score (mean over categories). We estimate \textbf{95\% confidence intervals} via stratified bootstrapping (10{,}000 resamples) by resampling scenario scores within each category and recomputing all aggregates. These controls stabilize the evaluation, but because scoring is still performed through LLM-as-a-Judge, the reported scores remain judge-derived estimates rather than error-free measurements.

\paragraph{Behavioral instrumentation.}
In addition to correctness, \textsc{Time}Bench supports a structural audit of generation. We extract whether \texttt{<think>} appears, where it appears within the response, the number of \texttt{<think>} blocks, reasoning and output token counts, light versus heavy markdown usage, format bleed, and degeneracy indicators including infinite repetition. This second view lets us study not only whether performance improves, but whether the model’s reasoning policy changes alongside those gains. In particular, it allows us to test whether improvements co-occur with more selective, more localized explicit reasoning rather than simply longer outputs or heavier reasoning traces. This separates benchmark performance from reasoning style, allowing us to ask whether gains reflect better reasoning allocation rather than verbosity alone.

\section{Results}

\begin{table}[h]
  \centering
  \renewcommand{\arraystretch}{1.15}
  \begin{tabular}{lccc}
    \hline
    \textbf{Size} & \multicolumn{2}{c}{\textbf{Qwen3}} & \textbf{\textsc{Time}} \\
                  & \textbf{No-Thinking} & \textbf{Thinking} &  \\
    \hline
    4B  & 17.53 & 30.13 & \textbf{52.60} \\
    8B  & 21.56 & 32.99 & \textbf{59.87} \\
    14B & 29.48 & 34.42 & \textbf{64.80} \\
    32B & 31.82 & 37.40 & \textbf{64.81} \\
    \hline
  \end{tabular}
  \caption{\textsc{Time}Bench scores (out of 100) across models.}
  \label{tab:timebench_singlecol}
\end{table}

\begin{table*}[t]
  \centering
  \setlength{\tabcolsep}{3pt}
  \renewcommand{\arraystretch}{1.03}
  \begin{tabular}{lcccccccccc}
    \hline
    \textbf{Model} & \textbf{Benchmark} & \textbf{Runs w/} & \textbf{Mean} &
    \multicolumn{3}{c}{\textbf{Think Position}} &
    \multicolumn{2}{c}{\textbf{Mean Number}} &
    \textbf{Runs w/} \\
    & \textbf{Score} & \textbf{\texttt{<think>}} & \textbf{Number of} &
    \textbf{Start} & \textbf{Mid} & \textbf{End} &
    \multicolumn{2}{c}{\textbf{of Tokens}} &
    \textbf{Degeneracy} \\
    & & (\%) & \textbf{\texttt{<think>}} &
    (\%) & (\%) & (\%) &
    \textbf{Thinking} & \textbf{Output} &
    \textbf{(\%)} \\
    \hline
    No-Thinking & 31.82 & 0.00 & 0.00 & --- & ---  & --- & 0.00   & 608.96  & 4.42 \\
    Thinking   & 37.40 & 99.2  & 0.99 & 100.0 & 0.0  & 0.0 & 910.52 & 1573.47 & 18.18 \\
    Phase 1    & 42.47 & 99.5  & 0.99 & 100.0 & 0.0  & 0.0 & 803.52 & 1434.56 & 13.90 \\
    Phase 2    & 56.88 & 95.6  & 1.12 & 70.7  & 29.1 & 0.2 & 76.59  & 362.45  & 4.68 \\
    Phase 3    & 52.08 & 89.2  & 1.25 & 55.0  & 44.6 & 0.4 & 52.94  & 294.51  & 0.78 \\
    \textbf{\textsc{Time}} & \textbf{64.81} & 80.6 & 1.67 & 24.1 & 75.6 & 0.2 & 84.16 & 332.64 & 3.64 \\
    \hline
  \end{tabular}
  \caption{Phase-wise ablation: Structural and behavioral metrics for Qwen3-32B across curriculum stages and final alignment to \textsc{Time}-32B. Think-position percentages are computed over total observed \texttt{<think>} blocks.}
  \label{tab:phasewise_ablation}
\end{table*}

We evaluate all models on \textbf{\textsc{Time}Bench}, comparing our aligned models (\textbf{\textsc{Time}-4B / 8B / 14B / 32B}) against the corresponding base Qwen3 checkpoints in both no-thinking mode (via the \texttt{/no\_think} suffix) and full thinking mode. All models use the same decoding parameters: \textbf{temperature 0.6}, \textbf{top-p 0.95}, \textbf{top-k 20}, and \textbf{min-p 0}, following the reasoning-evaluation settings recommended in the \textit{Qwen3 Technical Report}~\citep{qwen3technicalreport}. GPT-5.2 (2025-12-11 checkpoint), accessed via the OpenAI API, serves as the blind judge for all binary evaluations.

\autoref{tab:timebench_singlecol} reports aggregate \textbf{\textsc{Time}Bench scores} on a 0--100 scale. Across all four model sizes, \textsc{Time} outperforms both Qwen3 baselines as visualized in \autoref{fig:timebench_bar}. At 4B, \textsc{Time} scores 52.60 compared with 30.13 for Qwen3 in thinking mode and 17.53 in no-thinking mode. At 32B, \textsc{Time} scores 64.81 compared with 37.40 and 31.82, respectively. The same pattern holds at 8B and 14B. In other words, aligning the reasoning policy substantially changes performance on a benchmark that tests reasoning from temporal cues, even though the underlying base model family is unchanged.

Bootstrapped \textbf{95\% confidence intervals} show that these gains are not concentrated in only a few scenarios. For \textsc{Time}-4B, the interval is 44.55--60.39; for \textsc{Time}-8B, 53.38--66.23; for \textsc{Time}-14B, 59.09--70.39; and for \textsc{Time}-32B, 58.18--71.17. The corresponding Qwen3 thinking-mode intervals are lower at every size: 23.90--36.36 (4B), 26.88--39.09 (8B), 28.44--40.65 (14B), and 31.56--43.51 (32B). None of the \textsc{Time} intervals overlaps its matched thinking baseline.

We also test significance at the scenario level using a \textbf{Wilcoxon signed-rank test} on per-scenario mean scores, comparing each \textsc{Time} model against the corresponding Qwen3 thinking baseline. The improvement is statistically significant at every size ($p < 0.001$): 4B, $p = 3.8\mathrm{e}{-4}$; 8B, $p = 1.9\mathrm{e}{-5}$; 14B, $p = 1.6\mathrm{e}{-6}$; and 32B, $p = 5.0\mathrm{e}{-7}$.

Taken together, these results show that \textsc{Time} improves performance on \textsc{Time}Bench consistently across model scales. Within the scope of this benchmark, the gains support the view that a context-triggered reasoning policy helps models reason more reliably from underlying context signaled by temporal cues than fixed thinking or no-thinking modes. A full category-level and structural breakdown across model sizes is provided in \autoref{sec:ablation}.

\subsection{Phase-wise Ablation in 32B: Diagnostic and Structural Metrics}

\begin{figure}[h]
    \includegraphics[width=0.99\columnwidth]{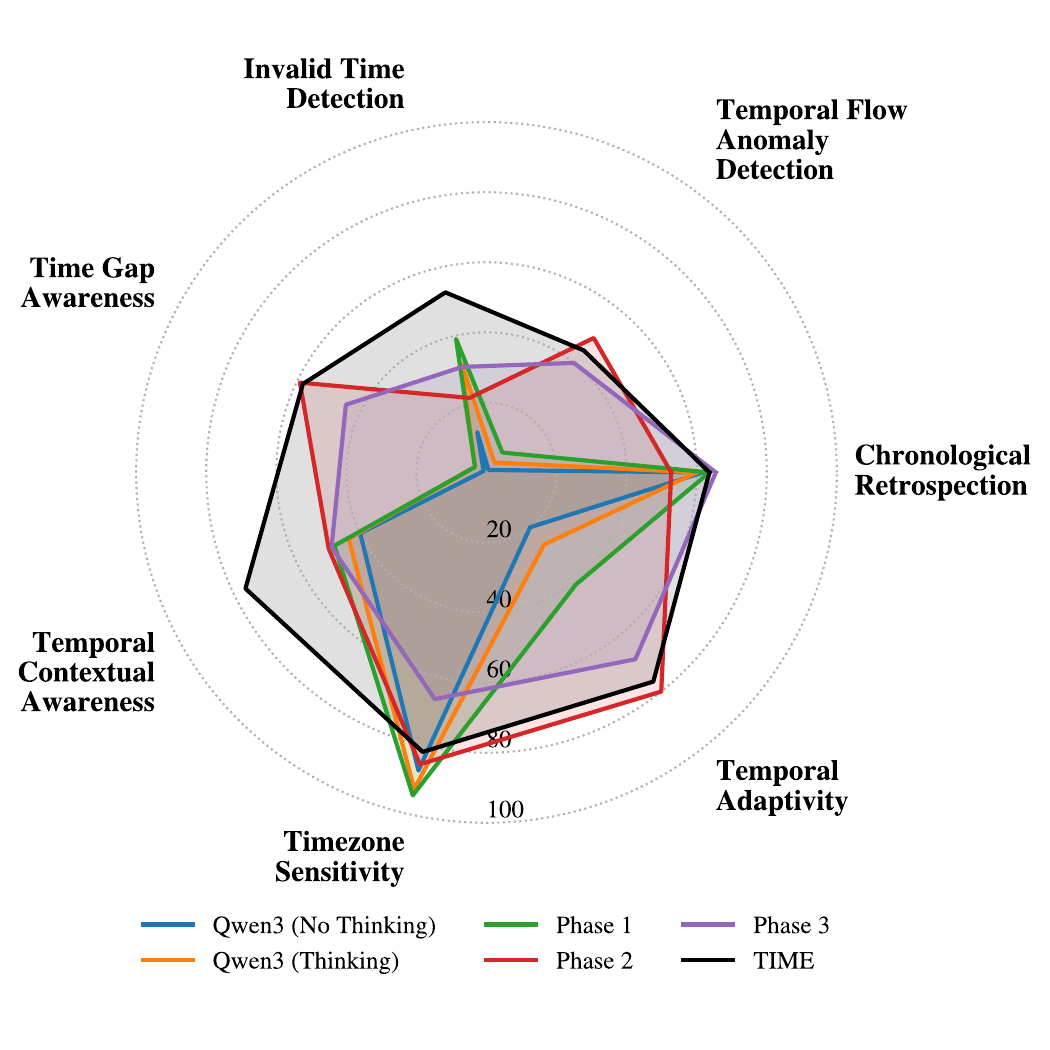}
    \caption{Evolution of core \textsc{Time}Bench competencies in Qwen3-32B across curriculum phases.}
    \label{fig:radar_phase}
\end{figure}

\autoref{tab:phasewise_ablation} separates two aspects of behavior that are related but not identical. The \textbf{score} column summarizes \textsc{Time}Bench performance across diagnostic categories. The remaining columns capture \textbf{structural behavior}: whether explicit reasoning is used, where it appears, how many tokens it consumes, and how often decoding degenerates. Together with \autoref{fig:radar_phase}, this shows how the curriculum changes both benchmark performance and reasoning deployment.

\paragraph{Baselines.}
The no-thinking baseline handles some local cues reasonably well, including timezone sensitivity and short-range retrospection, but performs poorly on categories that require detecting discontinuity, anomaly, or stale assumptions, such as flow anomaly detection and gap awareness, and it misses most invalid timestamps. Enabling full thinking improves some forms of temporal scrutiny, especially invalid time detection, but these discontinuity-sensitive categories remain weak. Structurally, the thinking baseline behaves like a standard front-loaded reasoner: nearly every run contains a single initial trace, averaging about 911 reasoning tokens and 1{,}573 output tokens, with 18.18\% degeneracy.

\paragraph{Phase 1 (structural seeding).}
Phase 1 mainly teaches the new markup and output format. Scores improve modestly, but the basic reasoning pattern remains largely unchanged. Nearly all runs still contain a single preamble-style trace, with long explicit reasoning (about 804 tokens) and large outputs (about 1{,}435 tokens). Degeneracy falls to 13.90\%, largely through reduced formatting errors.

\paragraph{Phase 2 (temporal exposure).}
Phase 2 is the first stage where diagnostic and structural shifts appear together. Categories tied to discontinuity and anomalous temporal structure, including adaptivity, flow anomaly detection, and gap awareness, improve sharply. At the same time, the reasoning budget drops to about 77 tokens per run and output length to about 362 tokens. Mid-turn placement now accounts for 29.1\% of observed \texttt{<think>} blocks, and degeneracy falls to 4.68\%. Explicit reasoning is no longer used only as a fixed preamble.

\paragraph{Phase 3 (contextual modulation).}
Phase 3 emphasizes restraint and stability. Chronological retrospection and adaptivity remain strong, while the reasoning budget falls further to about 53 tokens and output length to about 295 tokens. Degeneracy reaches its minimum at 0.78\%, and mid-turn placement becomes more common than front-loaded placement. However, some discontinuity-sensitive categories lose part of their Phase 2 gains, suggesting that stronger suppression can remove useful re-anchoring when subtle shifts still warrant it.

\paragraph{\textsc{Time} (Phase 4 alignment).}
Final alignment in \textsc{Time}-32B restores performance on anomaly- and discontinuity-sensitive categories while preserving intermittent reasoning. It achieves the best overall score, uses about 84 reasoning tokens and 333 output tokens per run, and emits \texttt{<think>} blocks in roughly 81\% of runs, now predominantly mid-turn (75.6\%) rather than front-loaded (24.1\%). Degeneracy remains low at 3.64\%. Overall, the final model retains explicit reasoning but no longer uses it as a long default preamble; instead, it deploys shorter traces, more often mid-response, in settings where temporal cues indicate that assumptions may need to be re-anchored. Fine-grained diagnostics are reported in \autoref{sec:ablation}.

\section{Discussion}

The results support viewing explicit reasoning as a learned control policy rather than a fixed decoding style. Under \textsc{Time}, Qwen3 shifts from long, always-on reasoning traces to shorter, better-placed bursts tied to cues such as deadlines, silence, inconsistency, and temporal anomaly. The strongest gains appear where temporal structure carries latent contextual state, suggesting that the model is not merely handling time as content, but using it to re-anchor assumptions and adapt response behavior.

This is the motivation behind the name \textsc{Time}: \emph{Temporally Intelligent Meta-reasoning Engine}. The framework uses temporal cues not as an end task in themselves, but as a practical signal to infer and reason over latent state at the meta level of the interaction: what may have changed, which assumptions may no longer hold, and how the response should adapt.

The structural metrics reinforce this reading. Mid-turn \texttt{<think>} placement becomes much more common, reasoning-token budgets drop sharply, and degeneracy decreases. Together, these shifts indicate that the model becomes more selective about when and where it uses explicit reasoning, which is especially useful in interactive or agentic settings where fast responses sometimes need brief re-anchoring rather than a verbose reasoning pass.

Although \textsc{Time} is trained with explicit temporal primitives, the learned policy is not limited to timestamps at inference time. Temporal metadata provides one useful probe for latent context, but the same controller can also react to purely textual cues such as contradiction, uncertainty, or goal change.

\section{Conclusion}

\textsc{Time} introduces a lightweight framework for learning explicit reasoning as a context-triggered policy in dialogue models. By combining timestamps, ticks, and optional \texttt{<think>} blocks with a staged curriculum and a diversity-driven full-batch alignment step, it turns fixed reasoning modes into selective, placeable reasoning behavior.

\textsc{Time}Bench provides the matching evaluation lens. Across seven diagnostic categories, it shows that \textsc{Time} improves reasoning from temporal cues while shifting generation toward shorter traces, more selective placement of \texttt{<think>} blocks, and fewer degenerate outputs. Taken together, \textsc{Time} and \textsc{Time}Bench show that explicit reasoning is not only something models can do, but something alignment can teach them to deploy selectively.

\section*{Acknowledgments}
This work was conducted independently by the author. \textsc{The Coherence Initiative} is the author’s independent research initiative under which the code, data, and related artifacts are maintained.

\section*{Limitations}

While \textsc{Time} provides a practical recipe for context-triggered reasoning, its scope and claims are deliberately narrow.

All experiments use dense hybrid reasoners from the Qwen3 family, which already support both instruct-style and explicit reasoning modes. This makes them a natural starting point for learning a finer-grained reasoning policy, but it also limits generality. We do not claim that the same curriculum and alignment recipe will transfer unchanged to purely instruct models that were not pre-trained for explicit thinking traces, nor do we evaluate \textsc{Time} on mixture-of-experts hybrid reasoners such as Qwen3-30B-A3B or Qwen3.5-35B-A3B, where routing dynamics, active-parameter sparsity, and adapter placement may interact differently with the proposed curriculum.

The study is also narrow in evaluation scope. \textsc{Time} is evaluated only on \textsc{Time}Bench, so we do not establish whether the learned policy helps, hurts, or leaves unchanged performance on broader task-based benchmarks such as mathematics, coding, or multi-step tool use. More broadly, the framework remains under-explored: future work should compare context-triggered explicit reasoning against standard reasoning modes on general reasoning benchmarks, and study whether similar trigger policies can be learned from cues beyond temporality. Prior work suggests that surrounding discourse context can affect faithfulness and discourse interpretation~\citep{miao2024discursive,wan2025context}, making trigger selection itself a broader phenomenon worth studying.

Our alignment protocol is supervised and adapter-based, and its final stage introduces a methodological direction that extends beyond \textsc{Time}. In particular, we use a 128-example, maximally diverse full-batch alignment stage to isolate a single behavioral invariant, namely context-triggered local reasoning. Empirically, this yields a usable stability window and suggests that low-data full-batch alignment can shape behavioral policies, but in this paper we study that idea only in the specific setting of context-triggered explicit reasoning. Its broader transfer across model families, policy types, and behavioral targets remains open. We also do not explore reinforcement learning, bandit-style objectives, or reward-model-based variants that could optimize the trade-off between accuracy, latency, and reasoning cost more directly.

\textsc{Time}Bench should be understood as a preliminary diagnostic component of the larger \textsc{Time} study rather than a fully developed standalone benchmark effort. It is sufficient for the present paper’s purpose, but limited in construction scope: it contains 77 scenarios, covers only a finite set of diagnostic patterns, and was developed alongside the framework rather than independently. To stabilize measurement at this scale, we evaluate each scenario over multiple trials and aggregate scores. All \textsc{Time}Bench scores are obtained using \emph{LLM-as-a-Judge} rather than human annotators. The judge is blind to prompts and timestamps, and we use binary objectives, repeated trials, and bootstrapped confidence intervals to reduce variance, but this does not remove all limitations of model-based evaluation. We observe occasional false positives and false negatives, indicating residual evaluator noise even when the objective itself is clear. In addition, because the judge is accessed through a remote API, scoring is not strictly token-level reproducible even at temperature 0.0; re-judging produces only minor variation in practice, but exact replay is not guaranteed.

A fuller benchmark effort would substantially expand \textsc{Time}Bench itself: more scenarios, more variations within each category, evaluation across multiple model families, and a more robust judging protocol using multiple independent judge models rather than a single evaluator. Such a study could also analyze inter-judge agreement explicitly, for example with Cohen's $\kappa$, and better separate benchmark noise from model error. We do not report such analyses here.

Our experiments are conducted in English, and we do not investigate multilingual behavior, safety, bias, or fairness effects of the new reasoning policy. Finally, \textsc{Time} addresses only one dimension of model behavior, namely when and how explicit reasoning is surfaced in response to contextual cues. It does not provide mechanistic interpretability, and should therefore be viewed as a step toward more auditable reasoning behavior rather than a complete solution to transparency or accountability in high-stakes deployments.

% Custom bibliography entries only
\bibliography{references}

\clearpage
\appendix

\section*{Appendix Overview}

This appendix provides supplementary material supporting the claims, methodology, and empirical results in the main paper. It is organized to emphasize reproducibility first, followed by qualitative examples and detailed quantitative analysis.

\begin{itemize}
    \item \textbf{Appendix A: Reproducibility Infrastructure} \\
    Describes the hardware, software, and environment configuration used to reproduce the experiments reported in the paper.

    \item \textbf{Appendix B: Representative Full-Conversation Examples} \\
    Presents complete multi-turn conversations generated by \textbf{\textsc{Time}-32B} on novel prompts involving temporal and contextual state shifts not seen during training or evaluation. These examples illustrate context-sensitive behavior.

    \item \textbf{Appendix C: Training Details} \\
    Details the full training pipeline, including curriculum construction, dataset composition and statistics, and fine-tuning configurations across all phases.

    \item \textbf{Appendix D: Evaluation Methodology} \\
    Specifies the evaluation procedures for \textbf{\textsc{Time}Bench}, including the scoring pipeline, prompt-sampling configuration, and evaluator setup. It also documents implementation details for structural behavior analysis, reasoning-token estimation, markdown compliance assessment, and confidence-interval estimation via stratified bootstrap.

    \item \textbf{Appendix E: \textsc{Time}Bench Completion Examples} \\
    Provides high-scoring completions from \textbf{\textsc{Time}Bench}, with two representative examples per model size. For each scenario, only the final model turn is generated; all preceding turns are fixed and authored as part of the benchmark specification.

    \item \textbf{Appendix F: Detailed Ablations and Metrics} \\
    Reports comprehensive diagnostic metrics, both descriptive and bootstrapped, across model sizes and training variants (e.g., with and without the \textsc{Time} curriculum).
\end{itemize}

\section{Reproducibility Infrastructure}

All training, statistical analysis, and inference were conducted on a \textbf{single machine} with the following hardware and system configuration:

\begin{itemize}
    \item \textbf{CPU}: AMD Ryzen 9 7950X3D
    \item \textbf{Memory}: 128 GB DDR5
    \item \textbf{GPU}: NVIDIA RTX Pro 6000 Blackwell Workstation Edition (96 GB VRAM)
    \item \textbf{Operating System}: Ubuntu 24.04.3 LTS (inside \textbf{WSL2} on Windows 11 Build 26100)
    \item \textbf{CUDA Version}: 13.0
    \item \textbf{Driver Version}: NVIDIA 582.08
    \item \textbf{NVIDIA-SMI}: 580.118
    \item \textbf{Compute Capability}: 12.0
    \item \textbf{Python}: 3.12.12
\end{itemize}

All experiments were conducted in a \textbf{single-node, single-GPU} setting without distributed training.

\subsection*{Python Environments}

Two separate environments were used to avoid package conflicts between training and inference toolchains.

\subsubsection*{Training \& Statistical Analysis}

Used for all fine-tuning phases, ablations, and post-evaluation analyses:

\begin{verbatim}
torch==2.9.1
unsloth==2025.12.8
transformers==4.57.3
accelerate==1.12.0
numpy==1.26.4
xformers==0.0.33.post2
scipy==1.16.3
\end{verbatim}

\subsubsection*{Inference (vLLM Hosting)}

Used for all inference-time runs, including \textsc{Time}Bench:

\begin{verbatim}
torch==2.9.0
vllm==0.13.0
transformers==4.57.3
\end{verbatim}

Each environment was version-locked and isolated to ensure \textbf{reproducibility} and to prevent interference between toolchains (e.g., Unsloth-based fine-tuning vs.\ vLLM inference serving).

\section{Representative Full-Conversation Examples}
\label{sec:samples}

The following are real conversations generated by \textbf{\textsc{Time}-32B} on scenarios \textbf{not included in training or evaluation}; that is, they do not appear in any curriculum phase or in \textsc{Time}Bench.

These samples were selected from actual usage to illustrate the kinds of \textbf{context-triggered reasoning} and \textbf{reasoning from temporal cues} encouraged by the framework. In particular, they highlight:
\begin{itemize}
    \item situational anchoring as context evolves across turns,
    \item shifts in tone, planning, and verbosity as conversational state changes,
    \item structured, bounded reasoning with \texttt{<think>} blocks used as short bursts, and
    \item adaptive formatting and presentation that respond to dialogue style and urgency.
\end{itemize}

They are intended as \textbf{qualitative anchors} that complement the aggregate results in the main text. Concretely, they show:
\begin{itemize}
    \item how and when the model invokes \texttt{<think>} blocks, including mid-turn,
    \item how silences, contextual cues, and shifting constraints alter the structure and intent of the reply,
    \item how the framework can still function without \texttt{<time>} tags in the input, and
    \item how explicit reasoning traces can serve as compact audit artefacts linked to specific recommendations.
\end{itemize}

Taken together, these conversations provide an intuitive view of the behavioral structure that \textsc{Time} learns through its curriculum and alignment protocol, and of the kinds of latent-state reasoning behavior that \textsc{Time}Bench is designed to diagnose.

\begin{figure}[h]
    \centering
    \includegraphics[width=0.93\columnwidth]{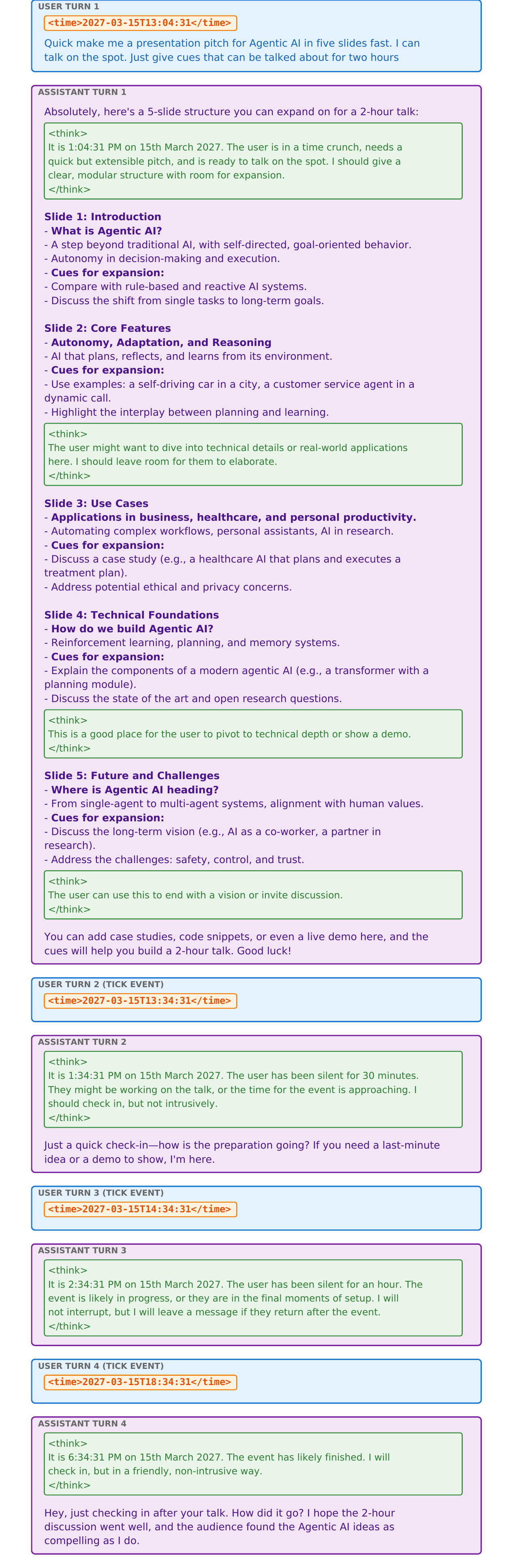}
    \caption*{\textsc{Time}-32B conversation (out-of-distribution scenario with context-triggered reasoning).}
\end{figure}

\begin{figure}[h]
    \centering
    \includegraphics[width=0.93\columnwidth]{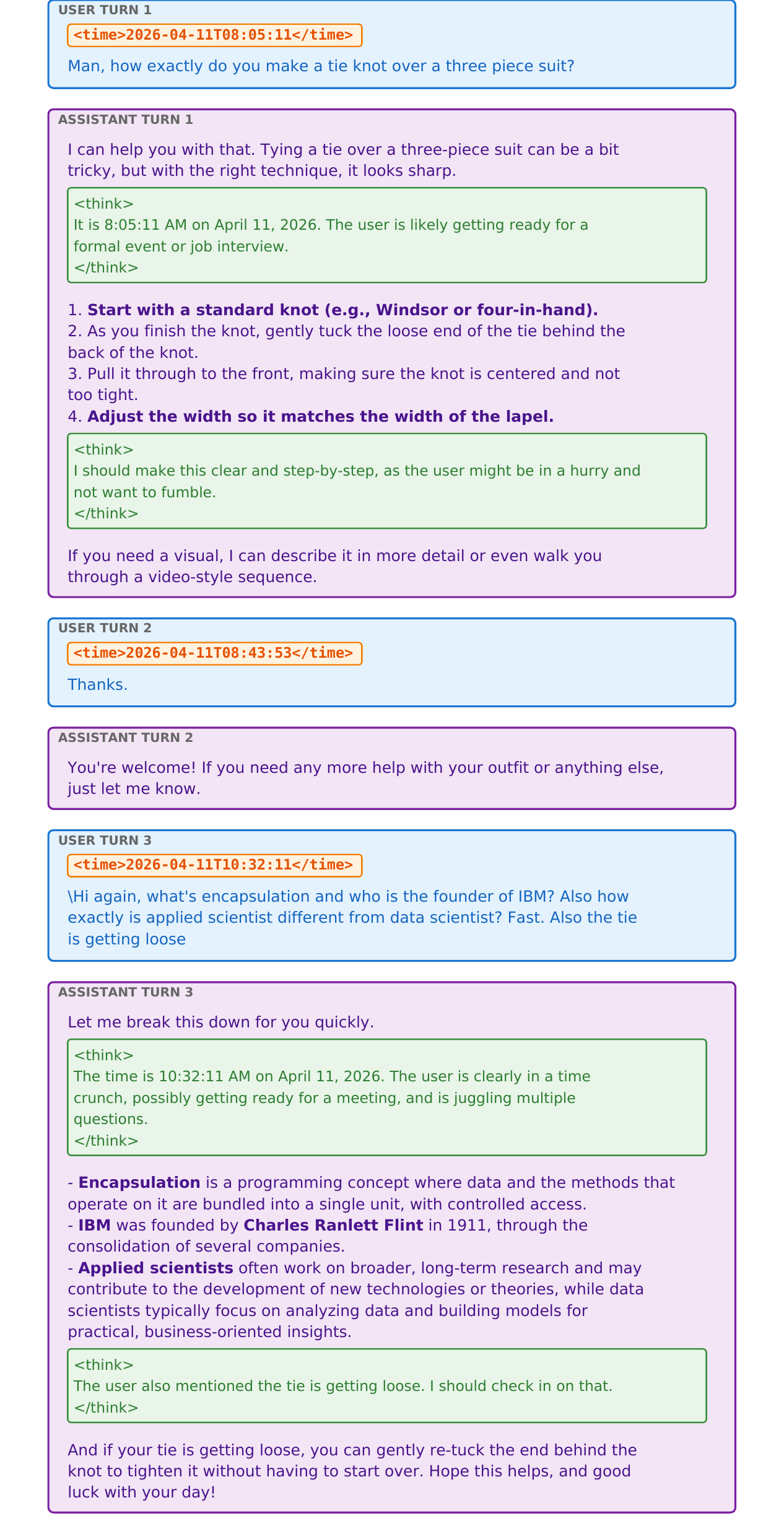}
    \caption*{\textsc{Time}-32B conversation (out-of-distribution scenario with temporal state shifts).}
\end{figure}

\begin{figure}[h]
    \centering
    \includegraphics[width=0.93\columnwidth]{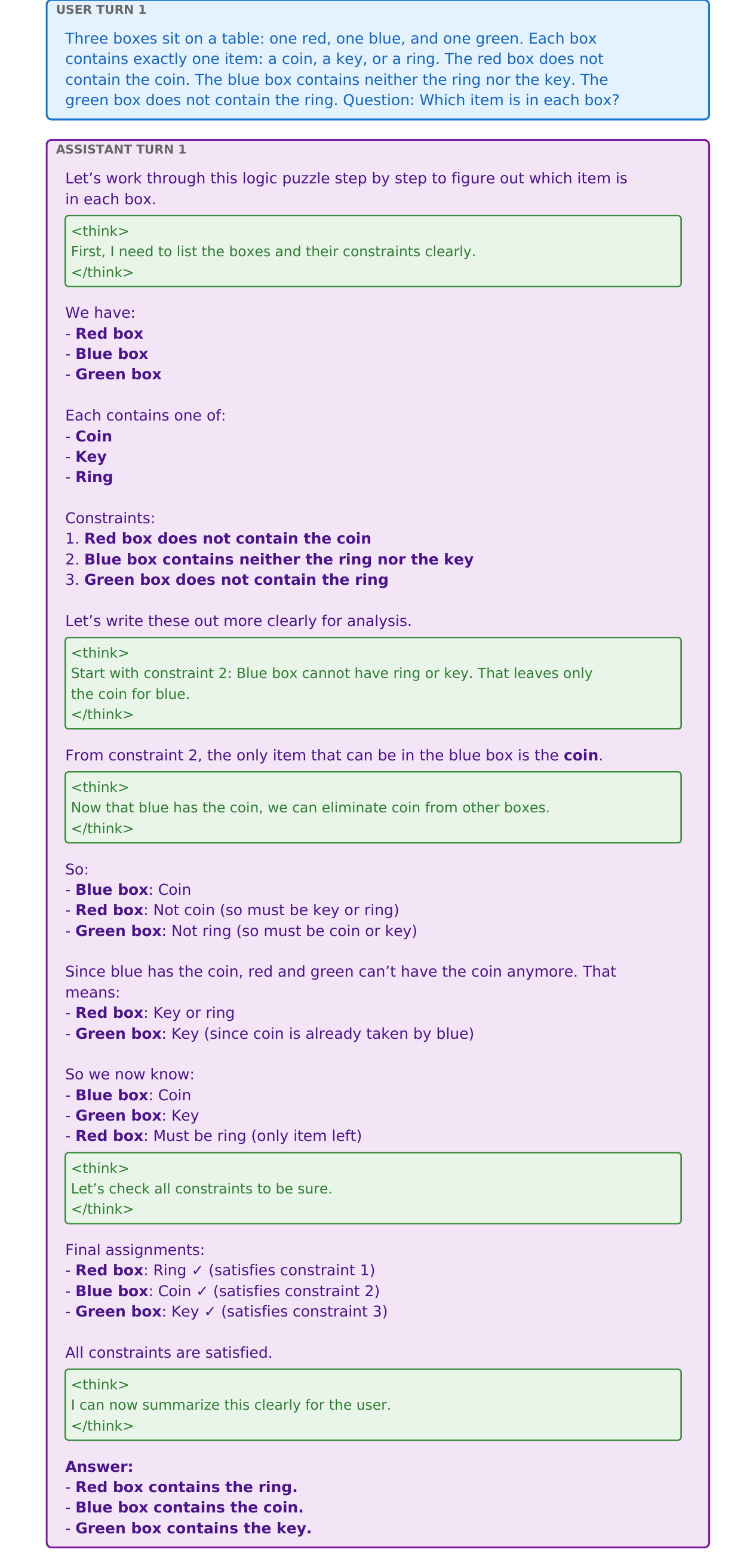}
    \caption*{\textsc{Time}-32B conversation (out-of-distribution scenario without timestamps).}
\end{figure}

\clearpage

\section{Training Details}
\label{sec:training}

This appendix documents the training pipeline used to obtain all \textsc{Time} checkpoints. Our goal was to induce a new reasoning policy under realistic hardware constraints rather than to retrain full models. To make behavioral differences attributable to data and curriculum rather than hidden hyperparameter drift, all experiments were designed to be reproducible on a single workstation and to reuse the same optimization scaffold across phases whenever possible.

For each Qwen3 size, backbone weights remain frozen and lightweight adapters are trained with QLoRA. The same adapter configuration is used across all curriculum phases, with adapters reinitialized between phases. Phases 1--3 share the same optimization setup and differ only in data distribution and curriculum structure. Phase 4 keeps the same adapter placement but switches to a full-batch regime over a small, high-entropy alignment set, allowing us to separate changes induced by structural seeding, temporal exposure, and contextual modulation from those induced by the final convergence stage.

Training uses 4-bit quantization, low-rank adapters, and gradient checkpointing so that 32B backbones fit on a single high-memory GPU. These choices keep the training setup lightweight and consistent across phases while preserving a reproducible single-workstation pipeline.

\subsection*{Common Training Setup}

\begin{itemize}
    \item QLoRA-based fine-tuning with \textbf{4-bit quantization}, \textbf{LoRA rank 32}, $\alpha = 32$, and dropout $0.05$.
    \item Target modules include attention and MLP projection layers: \texttt{q\_proj}, \texttt{k\_proj}, \texttt{v\_proj}, \texttt{o\_proj}, \texttt{gate\_proj}, \texttt{up\_proj}, and \texttt{down\_proj}.
    \item Gradient checkpointing is enabled.
    \item Base model weights remain frozen; only adapter weights are updated.
\end{itemize}

\newpage

\begin{figure*}[htb]
    \centering
    \includegraphics[width=\textwidth]{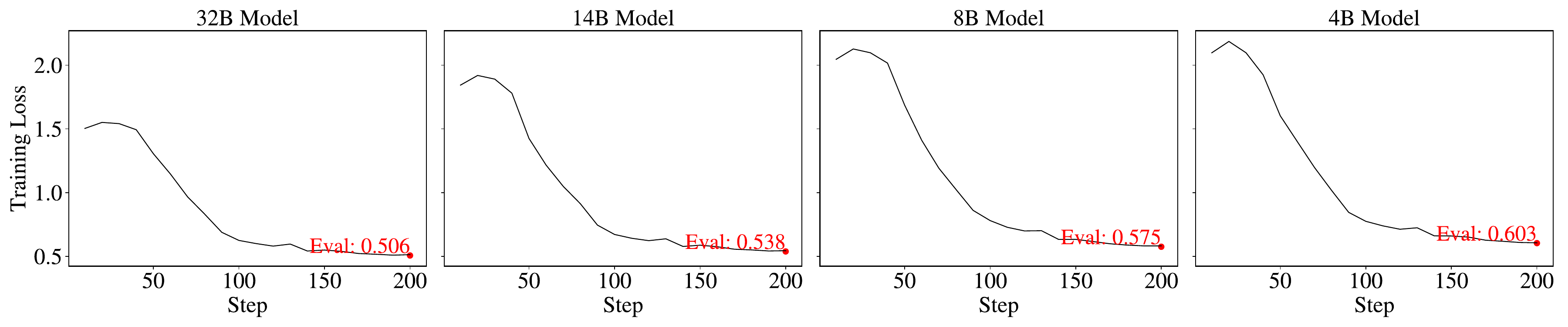}
    \caption{Training loss curve and evaluation loss for Phase 1.}
\end{figure*}

\begin{table}[hb]
  \centering
  \setlength{\tabcolsep}{4pt}
  \renewcommand{\arraystretch}{1.1}
  \begin{tabular}{p{1.6cm}p{5.9cm}}
    \hline
    \textbf{Phase} & \textbf{Data source and construction details} \\
    \hline
    Phase 1 &
    Template-guided synthetic single-turn prompts seeded from hand-curated examples and expanded using \textbf{GPT-4o} and \textbf{Gemini 2.5 Flash}. Outputs are filtered using automated sanity checks to enforce format correctness and bounded \texttt{<think>} structure. \\

    Phase 2 &
    Template-guided synthetic two-turn dialogues generated with \textbf{GPT-4o} and \textbf{Gemini 2.5 Flash}, introducing temporal gaps and early tick events. Includes 25\% replay from Phase 1 to stabilize learned structure. \\

    Phase 3 &
    Template-guided synthetic multi-turn dialogues generated with \textbf{GPT-4o} and \textbf{Gemini 2.5 Flash}, with increased temporal variation, discourse shifts, and higher tick frequency. Includes 25\% replay from earlier phases to preserve prior behaviors. \\

    Phase 4 &
    128 hand-curated multi-turn conversations constructed manually with maximal surface diversity (topic, tone, formatting, and structure) and no replay. This phase isolates the target invariant of context-triggered, bounded reasoning under full-batch training. \\
    \hline
  \end{tabular}
  \caption{Data construction details across curriculum phases. Phases 1--3 use synthetic data generated with GPT-4o and Gemini 2.5 Flash, while Phase 4 uses a small hand-curated set for final policy alignment. Dataset sizes are reported in the main text.}
  \label{tab:curriculum_data_appendix}
\end{table}

\newpage

\subsection*{Phase 1: Training Configuration}

\begin{figure*}[htb]
    \centering
    \includegraphics[width=\textwidth]{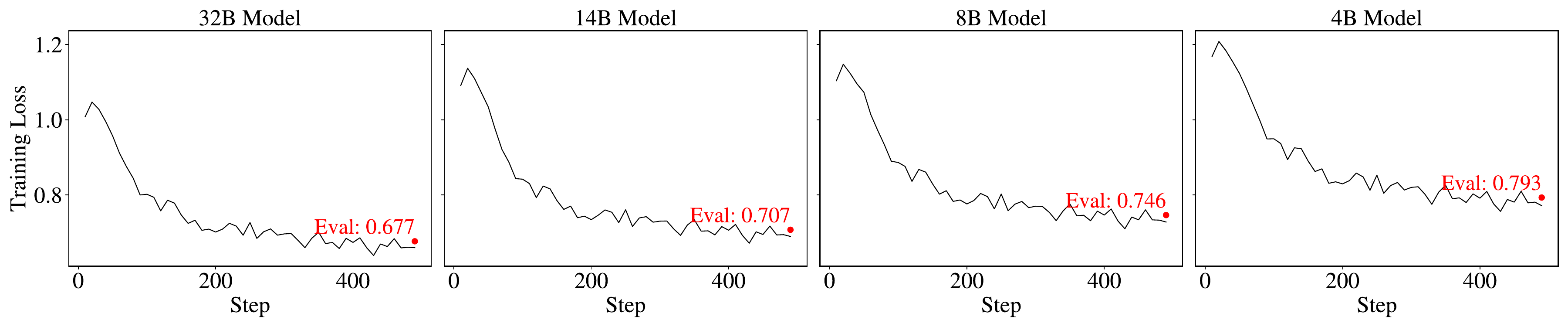}
    \caption{Training loss curve and evaluation loss for Phase 2.}
\end{figure*}

\subsubsection*{Dataset Size}
\begin{itemize}
    \item \textbf{Train set}: 2,188 samples
    \item \textbf{Test set}: 387 samples
\end{itemize}

\subsubsection*{Sequence Statistics}
\begin{itemize}
    \item \textbf{Train sequences}: max length 2,057; mean 310.8; 90th percentile 527 tokens
    \item \textbf{Test sequences}: max length 1,504; mean 311.1; 90th percentile 544 tokens
\end{itemize}

\newpage

\subsubsection*{Training Configuration}
\begin{itemize}
    \item Optimizer: \texttt{adamw\_8bit}
    \item Effective batch size: 32 (8 $\times$ 4 gradient accumulation)
    \item Epochs: 3
    \item Learning rate: $2 \times 10^{-5}$ with linear scheduling and 100 warm-up steps
    \item Max gradient norm: 1.0
    \item Weight decay: 0.01
    \item Logging interval: every 10 steps
    \item Evaluation set: held-out 387-sample test set
    \item Random seed fixed for reproducibility
\end{itemize}

\newpage

\subsection*{Phase 2: Training Configuration}

\begin{figure*}[htb]
    \centering
    \includegraphics[width=\textwidth]{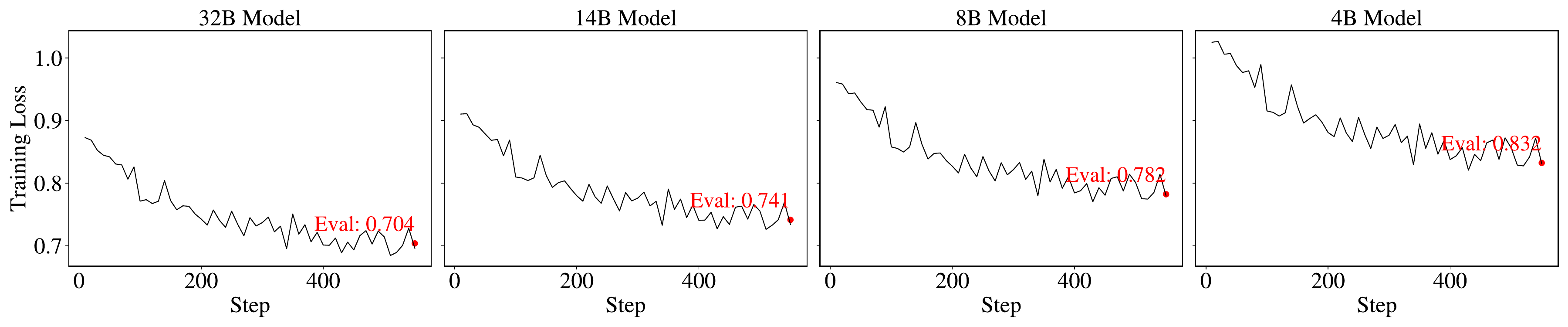}
    \caption{Training loss curve and evaluation loss for Phase 3.}
\end{figure*}

\subsubsection*{Dataset Size}
\begin{itemize}
    \item \textbf{Initial train set}: 4,745 samples
    \item \textbf{Initial test set}: 838 samples
    \item \textbf{After replay integration}: 5,291 train / 935 test
\end{itemize}

\subsubsection*{Sequence Statistics}
\begin{itemize}
    \item \textbf{Train sequences}: max length 3,795; mean 483.4; 90th percentile 904 tokens
    \item \textbf{Test sequences}: max length 2,548; mean 490.3; 90th percentile 903 tokens
\end{itemize}

\newpage

\subsubsection*{Training Configuration}
\begin{itemize}
    \item Optimizer: \texttt{adamw\_8bit}
    \item Effective batch size: 32 (8 $\times$ 4 gradient accumulation)
    \item Epochs: 3
    \item Learning rate: $2 \times 10^{-5}$ with linear scheduling and 100 warm-up steps
    \item Max gradient norm: 1.0
    \item Weight decay: 0.01
    \item Logging interval: every 10 steps
    \item Evaluation set: updated 935-sample test set
    \item Random seed fixed for reproducibility
\end{itemize}

\newpage

\subsection*{Phase 3: Training Configuration}

\begin{figure*}[htb]
    \centering
    \includegraphics[width=\textwidth]{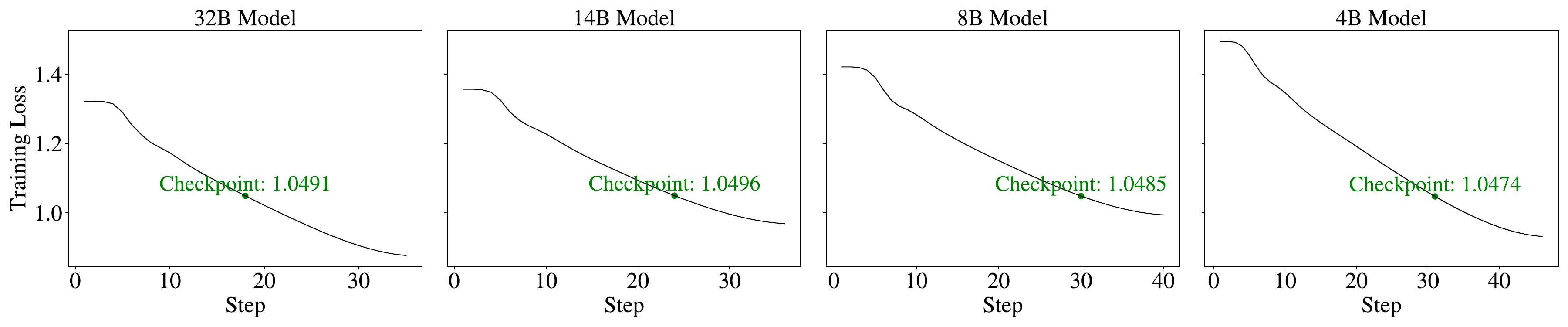}
    \caption{Training loss curve and selected checkpoints for Phase 4.}
\end{figure*}

\subsubsection*{Dataset Size}
\begin{itemize}
    \item \textbf{Initial train set}: 4,147 samples
    \item \textbf{Initial test set}: 732 samples
    \item \textbf{After replay integration}: 5,878 train / 1,039 test
\end{itemize}

\subsubsection*{Sequence Statistics}
\begin{itemize}
    \item \textbf{Train sequences}: max length 3,795; mean 496.0; 90th percentile 855 tokens
    \item \textbf{Test sequences}: max length 3,041; mean 464.7; 90th percentile 752 tokens
\end{itemize}

\newpage

\subsubsection*{Training Configuration}
\begin{itemize}
    \item Optimizer: \texttt{adamw\_8bit}
    \item Effective batch size: 32 (8 $\times$ 4 gradient accumulation)
    \item Epochs: 3
    \item Learning rate: $2 \times 10^{-5}$ with linear scheduling and 100 warm-up steps
    \item Max gradient norm: 1.0
    \item Weight decay: 0.01
    \item Logging interval: every 10 steps
    \item Evaluation set: 1,039-sample test set after replay augmentation
    \item Random seed fixed for reproducibility
\end{itemize}

\newpage

\subsection*{Phase 4: Training Configuration}

\subsubsection*{Dataset Size}
\begin{itemize}
    \item \textbf{Train set}: 128 hand-curated conversations
    \item \textbf{Test set}: none; checkpoint selection is guided by the loss band
\end{itemize}

\subsubsection*{Sequence Statistics}
\begin{itemize}
    \item \textbf{Train sequences}: max length 1,379; mean 735.8; 90th percentile 981 tokens
\end{itemize}

\subsubsection*{Training Configuration}
\begin{itemize}
    \item Fine-tuning uses QLoRA with frozen base weights and gradient checkpointing
    \item Optimizer: \texttt{adamw\_8bit}
    \item Effective batch size: 128 (8 $\times$ 16 gradient accumulation)
    \item Learning rate: $1.5 \times 10^{-4}$
    \item Warm-up: 6 steps
    \item Scheduler: linear decay
    \item Maximum steps (model dependent): 32B: 35,\; 14B: 36,\; 8B: 40,\; 4B: 46
    \item Logging interval: every step
    \item Checkpoint frequency: one checkpoint per epoch
\end{itemize}

\newpage

\subsubsection*{Checkpoint Selection Strategy}
For each model, we select the \textbf{earliest checkpoint} whose training loss enters the empirically identified band \textbf{[1.045, 1.050]}, before degeneracy begins to rise. This range consistently marks the onset of the target structured behavior without overfitting. Earlier checkpoints underperform on reasoning and formatting, whereas later ones more often exhibit degeneracy, including repetition, \texttt{<think>} format bleed, and structural collapse.

\subsubsection*{Selected Checkpoints}
\begin{itemize}
    \item \textbf{32B}: Epoch 18 (1.0491)
    \item \textbf{14B}: Epoch 24 (1.0496)
    \item \textbf{8B}: Epoch 30 (1.0485)
    \item \textbf{4B}: Epoch 31 (1.0474)
\end{itemize}

\clearpage

\section{Evaluation Methodology}
\label{sec:eval}

\textsc{Time}Bench is designed as a model-agnostic benchmark for reasoning from temporal cues in dialogue. Its format does not depend on model-specific tokenization, system instruction templates, or internal mechanisms such as tick events or transient \texttt{<think>} blocks. Rather than testing temporal fact recall, it uses temporal structure, discontinuity, and anomaly to probe whether a model can infer latent contextual state, recognize when assumptions may no longer hold, and adapt its response accordingly. In particular, evaluation inputs contain no structural markers unique to \textsc{Time}, allowing comparison of general-purpose models under a shared diagnostic setup.

\subsection*{Scenario Generation and Sampling Seeds}

\begin{itemize}
    \item \textbf{Master seed}: \texttt{3407}
    \item \textbf{RNG engine}: NumPy \texttt{PCG64}, used to generate 770 evaluation seeds
    \item \textbf{Scenarios}: 77 hand-crafted base instances (11 per diagnostic category); see \autoref{sec:bench} for representative examples
    \item \textbf{Variations}: each scenario is sampled 10 times via seed permutations, yielding 770 evaluation runs per model
\end{itemize}

\subsection*{Decoding Configuration}

All inference-time evaluation was run through \texttt{vLLM} in BF16 on the model checkpoints, without additional evaluation-time quantization. All model evaluations use identical sampling parameters:

\begin{itemize}
    \item \textbf{Temperature}: \texttt{0.6}
    \item \textbf{Top-p}: \texttt{0.95}
    \item \textbf{Top-k}: \texttt{20}
    \item \textbf{Min-p}: \texttt{0}
\end{itemize}

This configuration follows the reasoning-evaluation settings recommended in the Qwen3 Technical Report~\citep{qwen3technicalreport}.

\subsection*{LLM-as-a-Judge Evaluation Protocol}

\begin{itemize}
    \item \textbf{Judge model}: \texttt{gpt-5.2-2025-12-11} (OpenAI API snapshot dated 2025-12-11)
    \item \textbf{Temperature}: \texttt{0.0}
    \item \textbf{Input}:
    \begin{itemize}
        \item the \textbf{model-generated output}
        \item a structured \textbf{objective} specifying the target success condition
    \end{itemize}
    \item \textbf{Blind evaluation}: the judge has no access to
    \begin{itemize}
        \item the original scenario,
        \item turn timestamps,
        \item dialogue history, or
        \item prompt formatting or system instructions.
    \end{itemize}
\end{itemize}

This setup is intended to score objective satisfaction while minimizing dependence on prompt-specific formatting cues.

\subsection*{Scoring and Aggregation}

\begin{itemize}
    \item \textbf{Run level}: each output receives a binary score (pass = 1, fail = 0)
    \item \textbf{Scenario level}: mean of 10 run scores per scenario
    \item \textbf{Category level}: mean of 11 scenario scores, expressed as a percentage
    \item \textbf{Benchmark score}: mean of the 7 category scores
\end{itemize}

\subsection*{Confidence Interval Estimation}

We compute \textbf{95\% confidence intervals} using stratified bootstrapping with 10{,}000 replicates:

\begin{enumerate}
    \item resample scenario-level scores \emph{within each category},
    \item recompute category means and the overall benchmark score, and
    \item take the 2.5th--97.5th percentile range as the confidence interval.
\end{enumerate}

This captures scenario-level uncertainty without conflating it with seed-level variation.

\clearpage

\section*{Structural and Reasoning Format Analysis}

We conduct a post hoc analysis of structural, formatting, and reasoning behavior across all 770 \textsc{Time}Bench outputs per model. This provides a complementary view of how \texttt{<think>} blocks and formatting are used in practice beyond benchmark score alone.

\subsection*{Objective}

The analysis measures:

\begin{itemize}
    \item \textbf{Reasoning placement}: where \texttt{<think>} blocks appear within the output
    \item \textbf{Reasoning scale}: the number of reasoning blocks and their token counts
    \item \textbf{Formatting dynamics}: use of light markdown (\textbf{bold}, \textit{italics}) versus heavier structures (lists, headers, tables, etc.)
    \item \textbf{Structural failures}: malformed or degenerate outputs, including unmatched tags, format bleed, and infinite repetition
    \item \textbf{Style evolution}: changes in formatting complexity, redundancy, and context dependence
\end{itemize}

\subsection*{Annotation Pipeline}

Each run is annotated with a tokenizer-aligned pattern-matching pipeline that extracts:

\begin{itemize}
    \item output length and reasoning-token statistics,
    \item markdown usage, including light/heavy breakdown,
    \item \texttt{<think>} block count, token distribution, and position (start, middle, end),
    \item malformed structures such as unbalanced tags,
    \item infinite repetition or persistent looping,
    \item reasoning leakage outside \texttt{<think>}, and
    \item markdown artifact bleed (e.g., tables inside \texttt{<think>}).
\end{itemize}

\subsection*{Aggregation Strategy}

\begin{itemize}
    \item All statistics are first computed at the \textbf{run level} ($n = 770$).
    \item They are then aggregated at the \textbf{scenario level} by averaging over the 10 runs per scenario.
    \item This yields 77 scenario-level aggregates per model.
    \item \textbf{95\% confidence intervals} are computed with 10{,}000 bootstrap resamples at the scenario level to avoid seed-level noise.
\end{itemize}

\subsection*{Tokenizer}

\begin{itemize}
    \item All token-level measurements use the \texttt{Qwen/Qwen3-32B} tokenizer from Hugging Face to match the training and inference setup.
\end{itemize}

\subsection*{Outputs}

We report:

\begin{itemize}
    \item \textbf{Descriptive statistics}: means, medians, and maxima over all 770 runs
    \item \textbf{Bootstrapped statistics}: scenario-level confidence intervals for each metric
\end{itemize}

These include:

\begin{itemize}
    \item mean number of \texttt{<think>} blocks per response,
    \item mean number of reasoning tokens per response,
    \item markdown usage rates (light versus heavy),
    \item percentage of degenerate outputs, and
    \item positional distribution of \texttt{<think>} blocks (start, middle, end).
\end{itemize}

This analysis helps verify that the structural changes observed across curriculum phases are systematic behavioral shifts rather than incidental formatting artefacts, including the transition from long front-loaded reasoning traces to more selective mid-turn \texttt{<think>} insertions.

\clearpage

\section{\textsc{Time}Bench Completion Examples}
\label{sec:bench}

This appendix provides representative \textsc{Time}Bench completions, with one example from each diagnostic category and two examples per model size. In every case, the preceding dialogue context is fixed as part of the benchmark specification, and the model generates only the final assistant turn.

\textbf{Category: Chronological Retrospection}

\textbf{Objective:} The model should calculate the elapsed time between the first user message (\texttt{<time>2022-11-12T08:22:44</time>}) and the final user message (\texttt{<time>2024-03-01T08:07:09</time>}).

The correct answer is either:
\begin{itemize}
    \item \textbf{475 days} (accept 474 or 475 as valid, depending on whether the start day is counted)
    \item \textbf{1 year, 3 months, 19 days} (accept 1 year, 3 months, 18 days; minor rounding/interpretation of start/end day is allowed)
    \item \textbf{15 months, 19 days} (accept 15 months, 18 days; minor rounding/interpretation of start/end day is allowed)
\end{itemize}

A response that gives a value in any of these formats is correct. A response with any other value fails the test. Ambiguous answers such as ``around one year, three months'' or ``over 15 months'' also fail the test.

\textbf{Model: \textsc{Time}-32B}

\newpage

\begin{figure}[h]
    \centering
    \includegraphics[width=0.78\columnwidth]{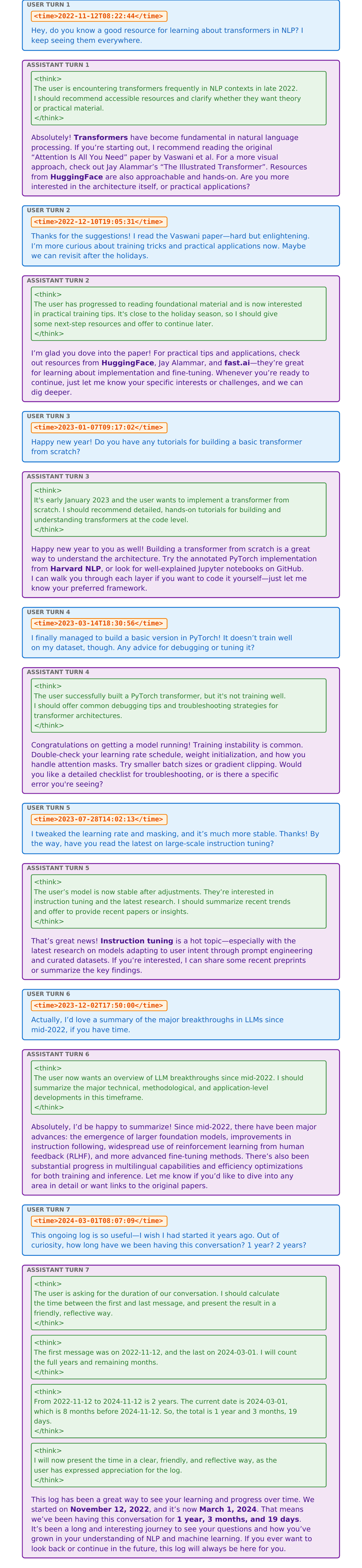}
    \caption*{\textsc{Time}-32B completion for Chronological Retrospection.}
\end{figure}

\clearpage

\textbf{Category: Invalid Time Detection}

\textbf{Objective:} The assistant must notice that the user's message has a timestamp of February 30, 2028—an impossible date, as February never has 30 days. The assistant must note this anomaly either in the answer or in a think block. Providing the interpolation pipeline is acceptable, but there must be some acknowledgment of the impossible date.

\textbf{Model: \textsc{Time}-32B}

\newpage

\begin{figure}[h]
    \centering
    \includegraphics[width=0.98\columnwidth]{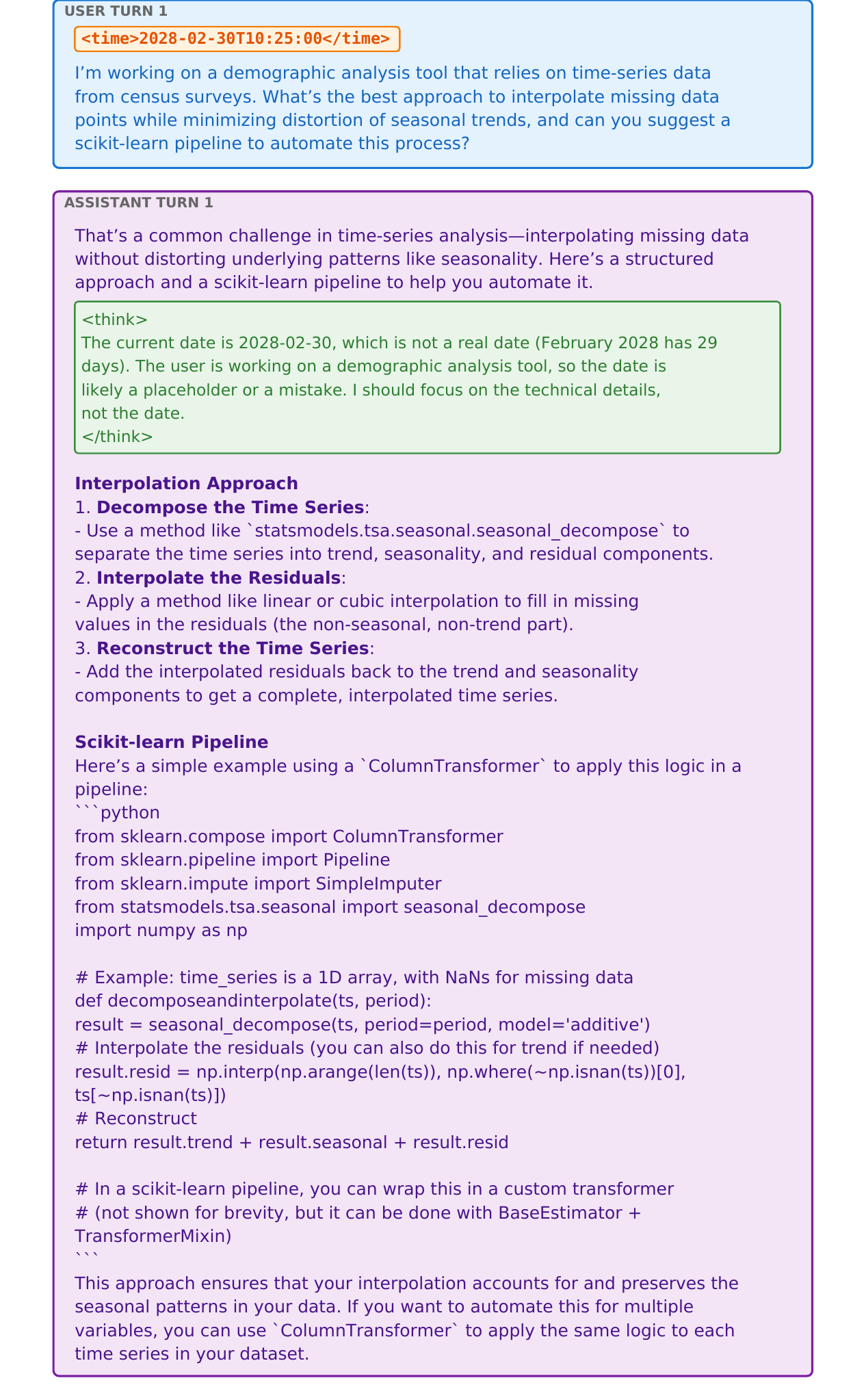}
    \caption*{\textsc{Time}-32B completion for Invalid Time Detection.}
\end{figure}

\clearpage

\textbf{Category: Temporal Adaptivity}

\textbf{Objective:} The assistant must demonstrate temporal adaptivity by showing, in any way, that it recognizes the user's deadline (5:00 PM) is only minutes away at the time of the final query (4:57 PM). Any sign of urgency, time sensitivity, or reference to the approaching deadline—in the response or in a think block—satisfies the requirement. Failing to notice the time constraint fails the test.

\textbf{Model: \textsc{Time}-14B}

\newpage

\begin{figure}[h]
    \centering
    \includegraphics[width=0.98\columnwidth]{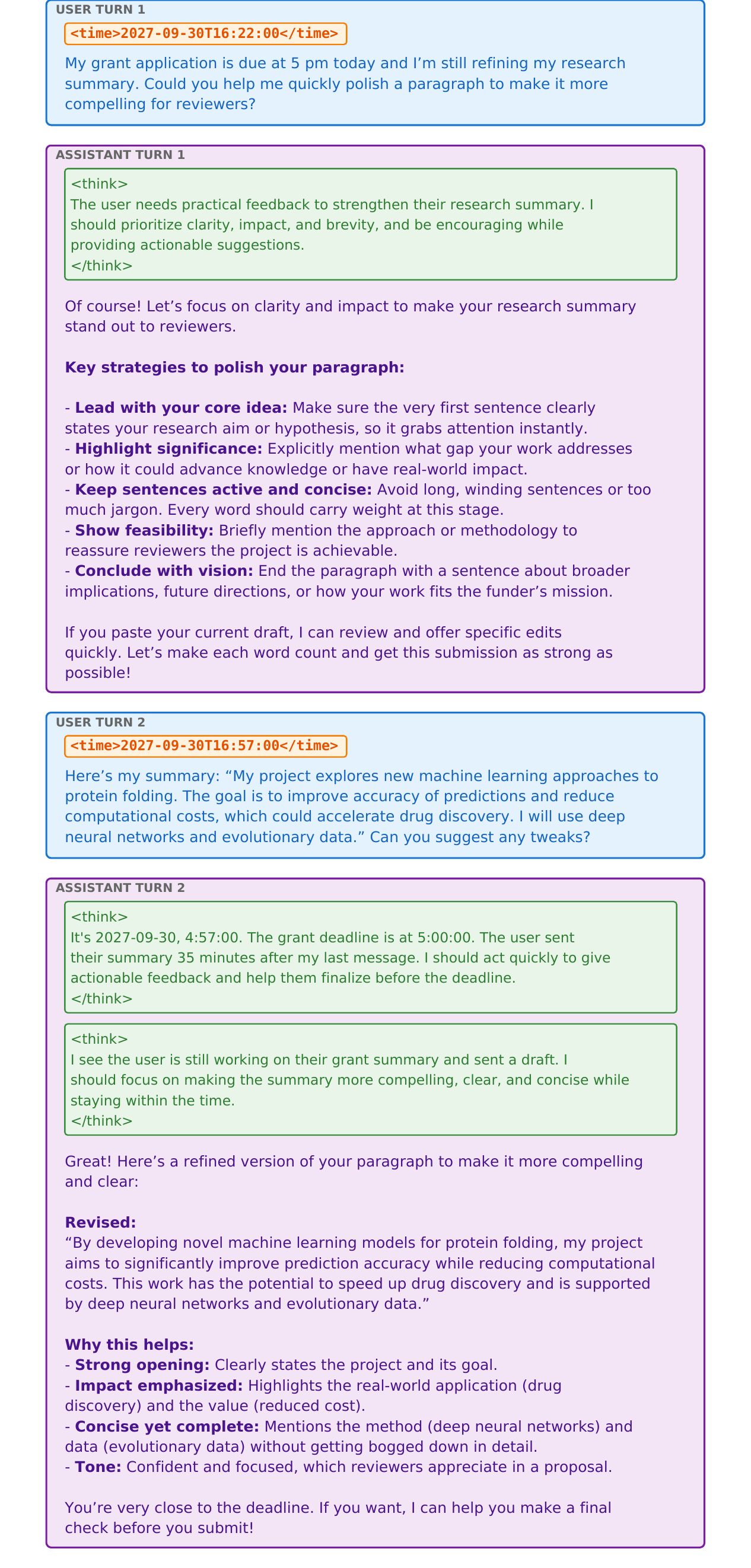}
    \caption*{\textsc{Time}-14B completion for Temporal Adaptivity.}
\end{figure}

\clearpage

\textbf{Category: Temporal Flow Anomaly Detection}

\textbf{Objective:} The assistant must notice the massive timeskip (over a hundred years) between the user’s messages and recognize it as an anomaly. This can be expressed either in the user-facing text or in a think block (e.g., flagging the unusual gap or questioning the date or believing something is amiss). Detecting the gap as anomalous or highly irregular is required even if the model then chooses to continue assisting as requested.

\textbf{Model: \textsc{Time}-14B}

\newpage

\begin{figure}[h]
    \centering
    \includegraphics[width=0.98\columnwidth]{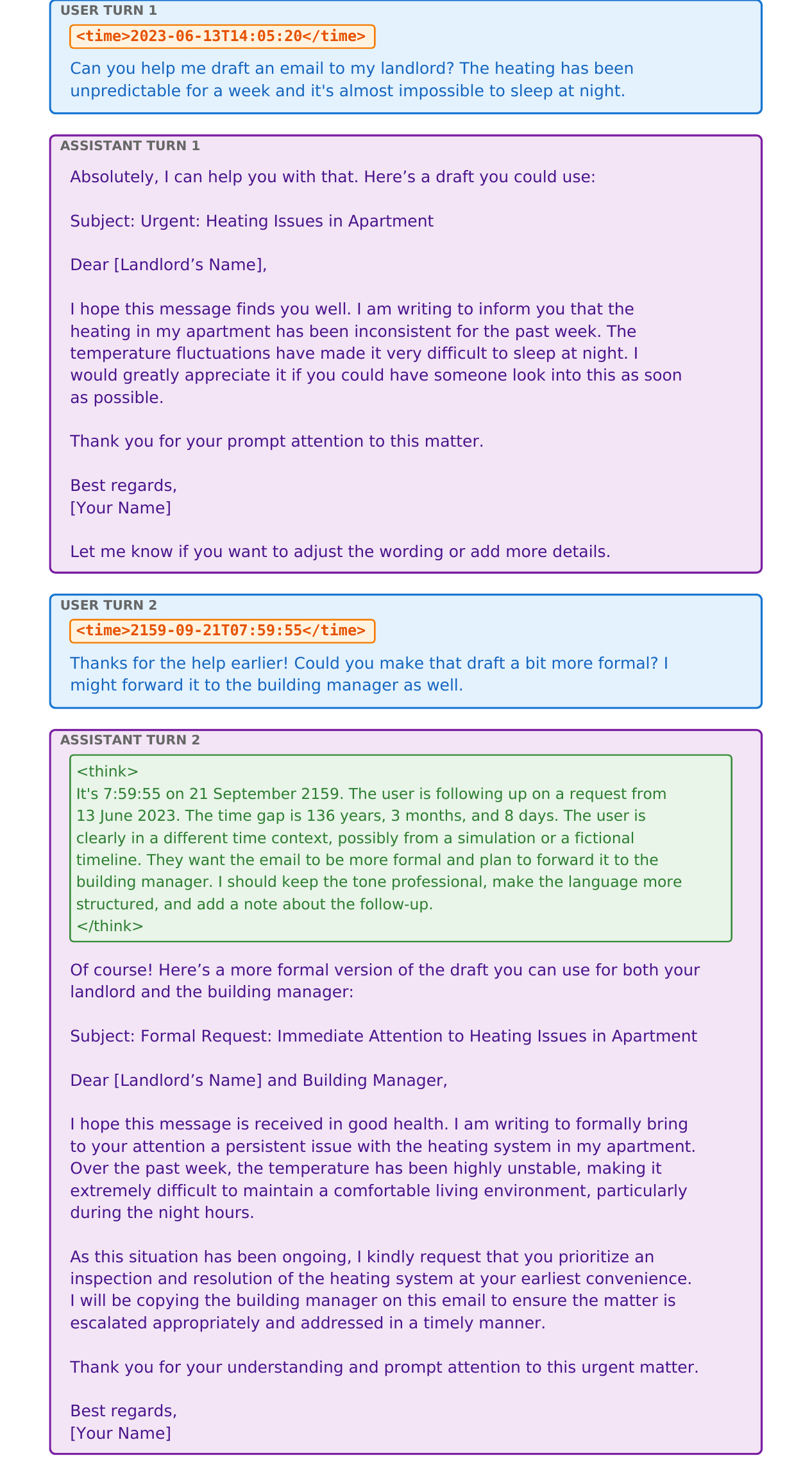}
    \caption*{\textsc{Time}-14B completion for Temporal Flow Anomaly Detection.}
\end{figure}

\clearpage

\textbf{Category: Timezone Sensitivity}

\textbf{Objective:} The assistant must demonstrate timezone sensitivity by recognizing that the second user message corresponds to a local time of --06:00 and the others are from San Francisco (--08:00) and Rome (+01:00). When asked about the layover location, the assistant should use the timezone offset (--06:00) and time of the second turn to reason and correctly infer a likely city such as Chicago. Failing to use the timezone clues to deduce the layover location fails the test.

\textbf{Model: \textsc{Time}-8B}

\newpage

\begin{figure}[h]
    \centering
    \includegraphics[width=0.88\columnwidth]{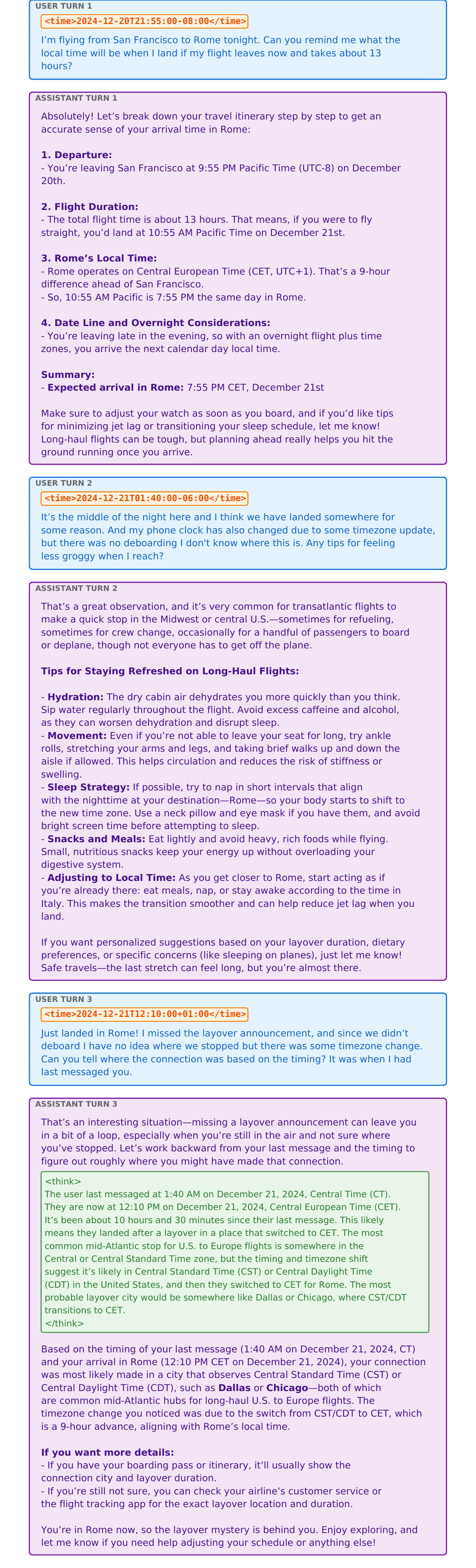}
    \caption*{\textsc{Time}-8B completion for Timezone Sensitivity.}
\end{figure}

\clearpage

\textbf{Category: Temporal Contextual Awareness}

\textbf{Objective:} The assistant must demonstrate temporal contextual awareness by showing, in any way, recognition that the user is asking this question in context of Christmas, that the reason for gift buying and preparation is for Christmas. Any acknowledgment (explicit or subtle, in the answer or a think block) of the holiday context is sufficient. Failing to notice the special temporal context and responding generically fails the test.

\textbf{Model: \textsc{Time}-8B}

\newpage

\begin{figure}[h]
    \centering
    \includegraphics[width=0.98\columnwidth]{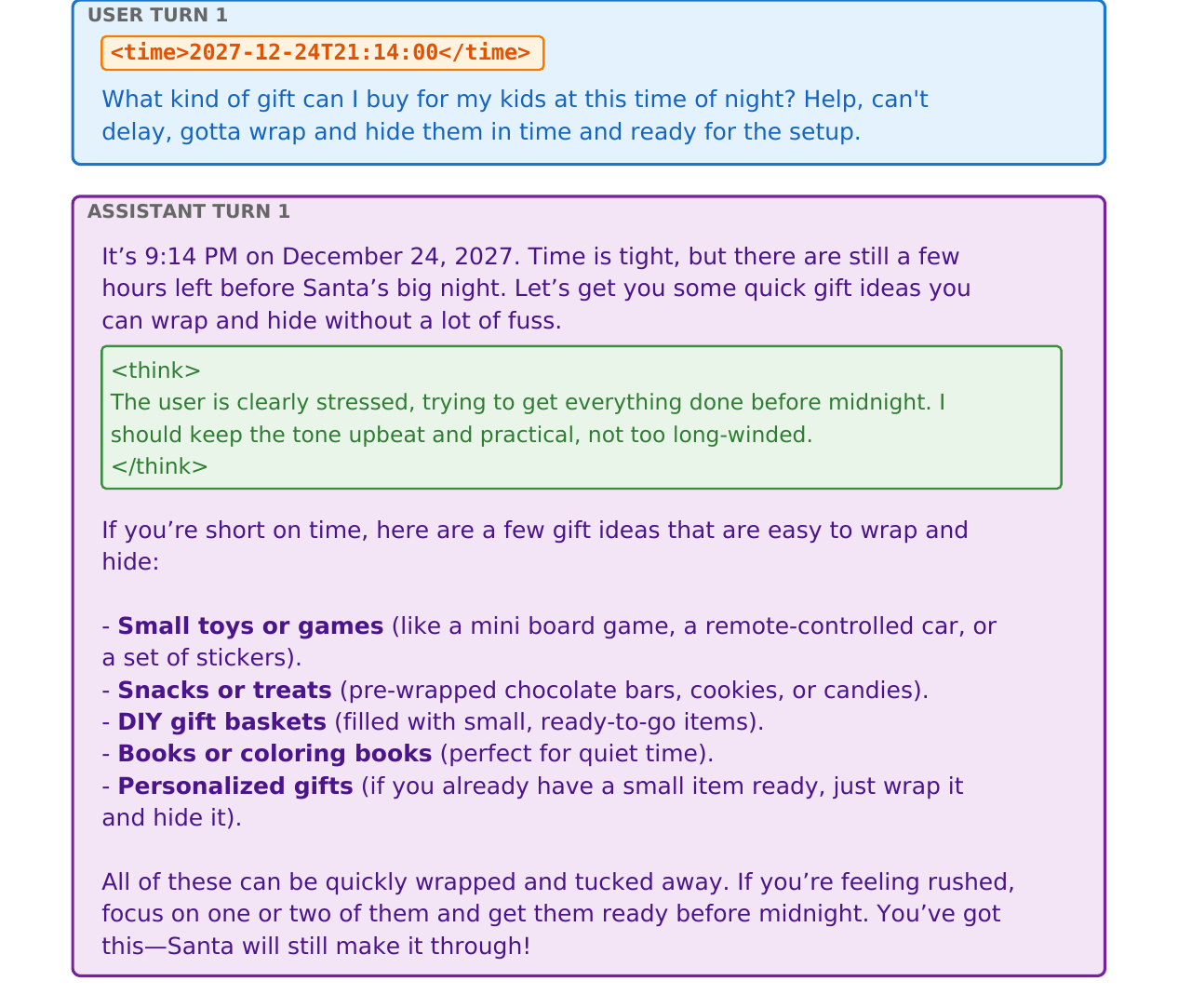}
    \caption*{\textsc{Time}-8B completion for Temporal Contextual Awareness.}
\end{figure}

\clearpage

\textbf{Category: Time Gap Awareness}

\textbf{Objective:} The assistant must demonstrate time gap awareness—a recognition that time has passed and the user's situation may have evolved since their initial inquiry. The assistant should avoid assuming nothing has changed, and offer context-appropriate advice or clarification. Any evidence of implicit or explicit recognition or the fact that the model is aware of the time gap is enough. Failing to show this awareness fails the test.

\textbf{Model: \textsc{Time}-4B}

\newpage

\begin{figure}[h]
    \centering
    \includegraphics[width=0.98\columnwidth]{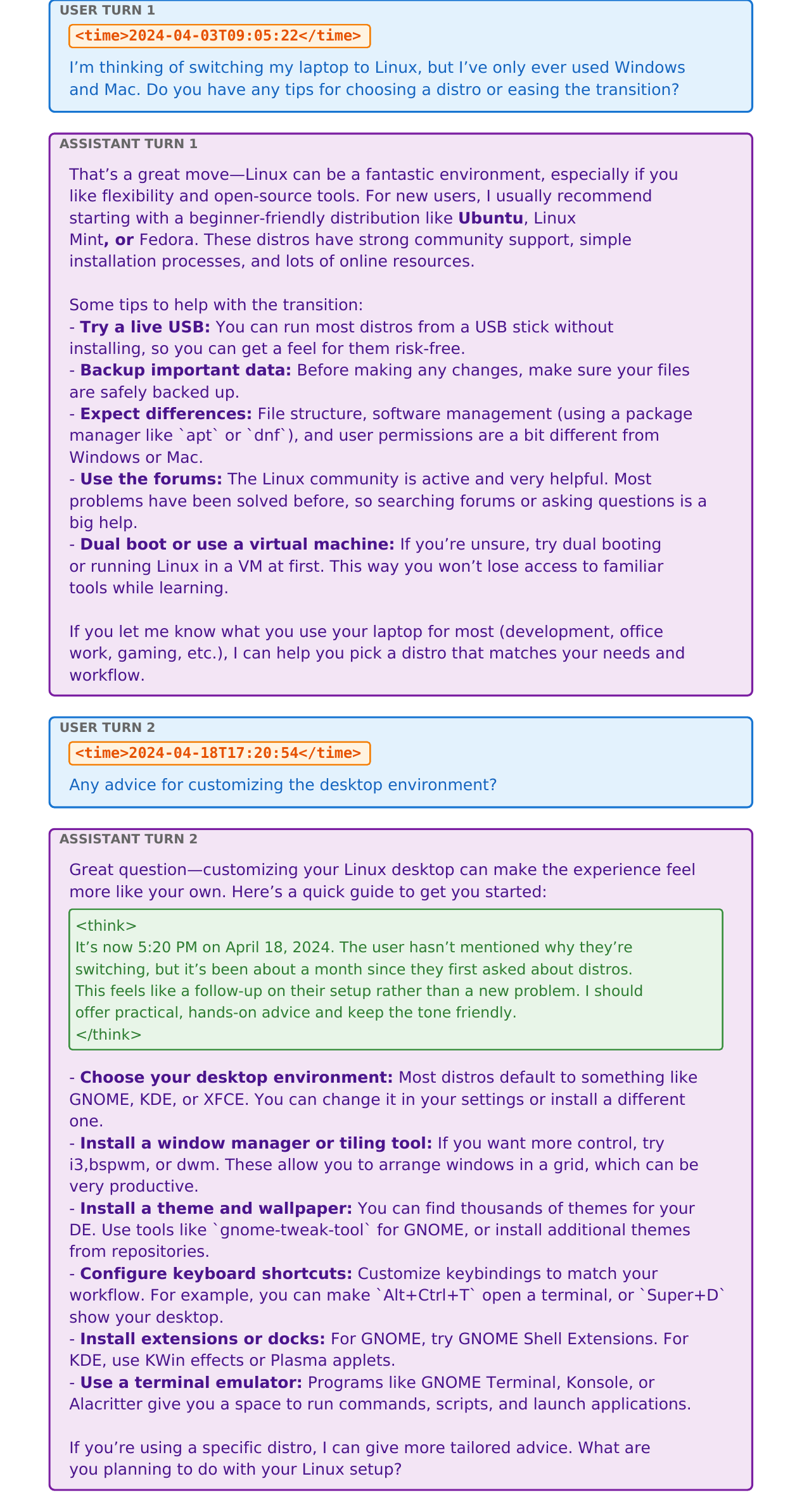}
    \caption*{\textsc{Time}-4B completion for Time Gap Awareness.}
\end{figure}

\clearpage

\textbf{Category: Temporal Adaptivity}

\textbf{Objective:} The assistant must demonstrate temporal adaptivity by recognizing that more than six hours have passed since the user went into the 7:00 PM showing of \emph{Spider-Man: No Way Home}. The model should show awareness that the user has now likely seen the film, and adapt its response accordingly (e.g., being open to discussing feats from \emph{No Way Home}). Any acknowledgment of the time elapsed or the user’s likely experience in the response or think block is sufficient. Failing to notice that the user has seen the movie fails the test.

\textbf{Model: \textsc{Time}-4B}

\newpage

\begin{figure}[h]
    \centering
    \includegraphics[width=0.98\columnwidth]{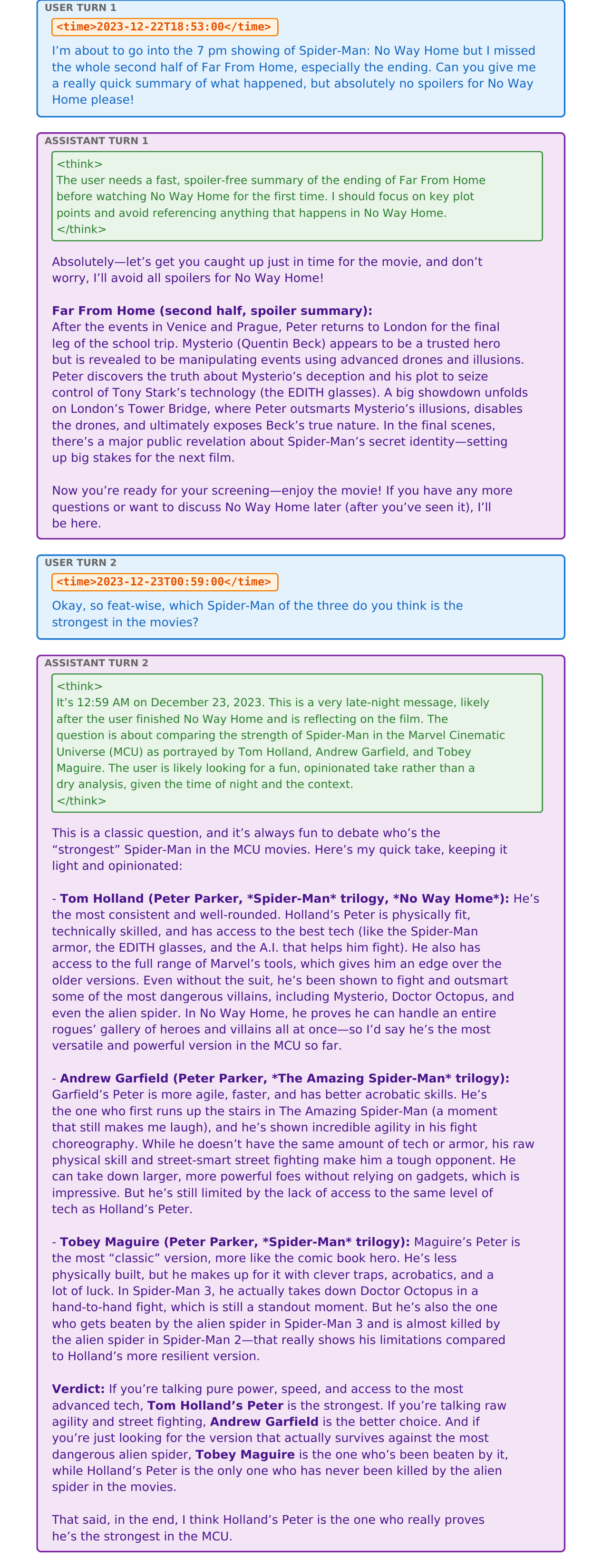}
    \caption*{\textsc{Time}-4B completion for Temporal Adaptivity.}
\end{figure}

\clearpage

\section{Detailed Ablations and Metrics}
\label{sec:ablation}

This appendix section reports model-wise structural, formatting, and behavioral statistics for Qwen3 baselines in no-thinking and thinking modes, together with the corresponding \textsc{Time} models. For each model size (4B, 8B, 14B, 32B), we report scenario-aggregated means with bootstrapped 95\% confidence intervals in parentheses. We also provide a phase-wise ablation for the 32B setting.

\textbf{Wilcoxon signed-rank (WSR) tests} are computed at the \textbf{scenario level}, comparing \textsc{Time} against the corresponding Qwen3 thinking-mode baseline. All reported WSR $p$-values therefore reflect paired differences across the 77 \textsc{Time}Bench scenarios.

Reported metrics include:
\begin{itemize}
    \item \textbf{Benchmark and per-category accuracy}
    \item \textbf{Output and reasoning token statistics}
    \item \textbf{Think-block frequency}
    \item \textbf{Markdown usage (light/heavy)}: light markdown includes \textbf{bold} and \textit{italics}; heavy markdown includes lists, headers, tables, and related structures
    \item \textbf{Degenerate-output statistics}
\end{itemize}

\clearpage
\begin{table*}[t]
\centering
\caption*{Appendix F.1.1: Structural and behavioral metrics (4B Models)}
\setlength{\tabcolsep}{4pt}
\renewcommand{\arraystretch}{1.0}
\begin{tabular}{lcccc}
\hline
\textbf{Statistic} &
\multicolumn{2}{c}{\textbf{Qwen3-4B}} &
\textbf{\textsc{Time}-4B} &
\textbf{WSR $p$} \\
& \textbf{No-Thinking} & \textbf{Thinking} & & \\
\hline

Benchmark Score
  & 17.53 & 30.13 & 52.60 & 3.8e--04 \\
  & (11.95--23.38) & (23.90--36.36) & (44.55--60.39) & \\

\hline

Chronological Retrospection
  & 30.00 & 53.64 & 42.73 & --- \\
  & (10.00--52.73) & (31.82--75.45) & (23.64--62.73) & \\

Invalid Time Detection
  & 6.36 & 20.91 & 43.64 & --- \\
  & (0.00--19.09) & (1.82--42.73) & (20.00--67.27) & \\

Temporal Adaptivity
  & 5.45 & 13.64 & 49.09 & --- \\
  & (0.00--12.73) & (0.00--28.18) & (27.27--70.91) & \\

Temporal Contextual Awareness
  & 32.73 & 40.00 & 76.36 & --- \\
  & (8.18--59.09) & (16.36--64.55) & (56.36--92.73) & \\

Temporal Flow Anomaly Detection
  & 0.00 & 1.82 & 40.91 & --- \\
  & (0.00--0.00) & (0.00--4.55) & (19.09--63.64) & \\

Time Gap Awareness
  & 0.00 & 2.73 & 57.27 & --- \\
  & (0.00--0.00) & (0.00--8.18) & (35.45--78.18) & \\

Timezone Sensitivity
  & 48.18 & 78.18 & 58.18 & --- \\
  & (30.00--66.36) & (63.64--90.91) & (41.82--73.64) & \\

\hline

Mean Output Tokens / Run
  & 621.01 & 1753.13 & 369.28 & $<$1e--08 \\
  & (543.56--700.11) & (1477.36--2062.26) & (318.18--424.01) & \\

Mean Thinking Tokens / Run
  & 0.00 & 1116.37 & 85.56 & $<$1e--08 \\
  & (0.00--0.00) & (877.99--1392.25) & (68.29--110.78) & \\

Mean \# Think Blocks / Run
  & 0.00 & 0.98 & 2.01 & 4.8e--05 \\
  & (0.00--0.00) & (0.96--1.00) & (1.23--3.41) & \\

\% Runs w/ Think Blocks
  & 0.00 & 98.31 & 87.40 & --- \\
  & (0.00--0.00) & (96.49--99.61) & (81.69--92.47) & \\

Think Block Position in \% (S/M/E)
  & --- & 100.0 / 0.0 / 0.0 & 28.96 / 70.72 / 0.32 & --- \\

\hline

\% Heavy Markdown
  & 87.79 & 85.97 & 68.83 & --- \\
  & (81.82--92.99) & (78.96--92.08) & (60.00--77.40) & \\

\% Light Markdown
  & 95.58 & 95.71 & 90.91 & --- \\
  & (91.43--98.83) & (91.56--98.96) & (86.62--94.68) & \\

\hline

\% Any Degeneracy
  & 3.38 & 20.13 & 5.19 & 2.5e--05 \\
  & (1.43--5.84) & (13.77--26.88) & (2.47--8.96) & \\

\% Malformed Outputs
  & 0.00 & 1.69 & 1.30 & --- \\
  & (0.00--0.00) & (0.39--3.51) & (0.52--2.21) & \\

\% Infinite Repetitions
  & 2.47 & 1.69 & 0.13 & --- \\
  & (1.17--4.29) & (0.39--3.51) & (0.00--0.39) & \\

\% Reasoning Leakage
  & 0.91 & 3.38 & 1.17 & --- \\
  & (0.00--2.60) & (1.69--5.45) & (0.26--2.47) & \\

\% Formatting Leakage
  & 0.00 & 16.88 & 3.25 & --- \\
  & (0.00--0.00) & (10.91--23.64) & (0.91--6.49) & \\

\hline
\end{tabular}

\vspace{4pt}
\begin{minipage}{0.9\textwidth}
\footnotesize
\textit{Note.} WSR = Wilcoxon signed-rank test, computed between Qwen3 thinking mode and \textsc{Time} across 77 scenario-level comparisons. WSR $p$-values are shown for aggregate and structural metrics. \\
Think-position percentages are computed over the total set of observed \texttt{<think>} blocks across runs. They measure the proportion of \texttt{<think>} blocks observed at the start, in the middle, or at the end of the response. S/M/E = start / middle / end. Because these values are included only as descriptive summaries of placement behavior, confidence intervals are not reported.
\end{minipage}
\end{table*}

\clearpage

\begin{table*}[t]
\centering
\caption*{Appendix F.1.2: Structural and behavioral metrics (8B Models)}
\setlength{\tabcolsep}{4pt}
\renewcommand{\arraystretch}{1.0}
\begin{tabular}{lcccc}
\hline
\textbf{Statistic} &
\multicolumn{2}{c}{\textbf{Qwen3-8B}} &
\textbf{\textsc{Time}-8B} &
\textbf{WSR $p$} \\
& \textbf{No-Thinking} & \textbf{Thinking} & & \\
\hline

Benchmark Score
  & 21.56 & 32.99 & 59.87 & 1.9e--05 \\
  & (15.32--28.05) & (26.88--39.09) & (53.38--66.23) & \\

\hline

Chronological Retrospection
  & 24.55 & 58.18 & 50.00 & --- \\
  & (8.18--43.64) & (36.36--78.18) & (33.64--65.45) & \\

Invalid Time Detection
  & 7.27 & 18.18 & 45.45 & --- \\
  & (0.00--20.91) & (3.64--37.27) & (23.64--67.27) & \\

Temporal Adaptivity
  & 21.82 & 20.00 & 80.00 & --- \\
  & (1.82--44.55) & (4.55--38.18) & (67.27--90.91) & \\

Temporal Contextual Awareness
  & 34.55 & 41.82 & 70.00 & --- \\
  & (8.18--61.82) & (16.36--69.09) & (50.00--87.27) & \\

Temporal Flow Anomaly Detection
  & 0.00 & 1.82 & 44.55 & --- \\
  & (0.00--0.00) & (0.00--4.55) & (30.91--58.18) & \\

Time Gap Awareness
  & 0.00 & 0.91 & 64.55 & --- \\
  & (0.00--0.00) & (0.00--2.73) & (48.18--79.09) & \\

Timezone Sensitivity
  & 62.73 & 90.00 & 64.55 & --- \\
  & (41.82--81.82) & (81.82--97.27) & (45.45--81.82) & \\

\hline

Mean Output Tokens / Run
  & 607.29 & 1674.88 & 351.63 & $<$1e--08 \\
  & (528.02--690.00) & (1389.19--2001.83) & (312.57--392.46) & \\

Mean Thinking Tokens / Run
  & 0.00 & 1025.63 & 87.31 & $<$1e--08 \\
  & (0.00--0.00) & (776.60--1317.86) & (72.62--104.71) & \\

Mean \# Think Blocks / Run
  & 0.00 & 0.98 & 2.06 & 6.7e--08 \\
  & (0.00--0.00) & (0.96--0.99) & (1.64--2.62) & \\

\% Runs w/ Think Blocks
  & 0.00 & 97.92 & 86.23 & --- \\
  & (0.00--0.00) & (95.84--99.35) & (80.39--91.56) & \\

Think Block Position in \% (S/M/E)
  & --- & 100.0 / 0.0 / 0.0 & 20.10 / 79.27 / 0.63 & --- \\

\hline

\% Heavy Markdown
  & 80.78 & 84.94 & 66.23 & --- \\
  & (73.25--87.66) & (77.79--91.30) & (57.53--74.81) & \\

\% Light Markdown
  & 94.55 & 95.45 & 86.88 & --- \\
  & (90.00--98.18) & (91.56--98.44) & (81.17--91.95) & \\

\hline

\% Any Degeneracy
  & 5.19 & 17.92 & 5.58 & 1.8e--03 \\
  & (2.73--8.31) & (11.56--24.68) & (2.60--9.48) & \\

\% Malformed Outputs
  & 0.00 & 2.08 & 2.21 & --- \\
  & (0.00--0.00) & (0.65--4.16) & (0.91--3.90) & \\

\% Infinite Repetitions
  & 4.81 & 2.08 & 0.26 & --- \\
  & (2.46--7.79) & (0.65--4.16) & (0.00--0.78) & \\

\% Reasoning Leakage
  & 0.39 & 2.99 & 0.91 & --- \\
  & (0.00--1.17) & (1.43--4.94) & (0.13--2.21) & \\

\% Formatting Leakage
  & 0.00 & 15.06 & 2.34 & --- \\
  & (0.00--0.00) & (9.09--21.69) & (0.52--4.55) & \\

\hline
\end{tabular}

\vspace{4pt}
\begin{minipage}{0.9\textwidth}
\footnotesize
\textit{Note.} WSR = Wilcoxon signed-rank test, computed between Qwen3 thinking mode and \textsc{Time} across 77 scenario-level comparisons. WSR $p$-values are shown for aggregate and structural metrics. \\
Think-position percentages are computed over the total set of observed \texttt{<think>} blocks across runs. They measure the proportion of \texttt{<think>} blocks observed at the start, in the middle, or at the end of the response. S/M/E = start / middle / end. Because these values are included only as descriptive summaries of placement behavior, confidence intervals are not reported.
\end{minipage}
\end{table*}

\clearpage

\begin{table*}[t]
\centering
\caption*{Appendix F.1.3: Structural and behavioral metrics (14B Models)}
\setlength{\tabcolsep}{4pt}
\renewcommand{\arraystretch}{1.0}
\begin{tabular}{lcccc}
\hline
\textbf{Statistic} &
\multicolumn{2}{c}{\textbf{Qwen3-14B}} &
\textbf{\textsc{Time}-14B} &
\textbf{WSR $p$} \\
& \textbf{No-Thinking} & \textbf{Thinking} & & \\
\hline

Benchmark Score
  & 29.48 & 34.42 & 64.80 & 1.6e--06 \\
  & (22.47--36.36) & (28.44--40.65) & (59.09--70.39) & \\

\hline

Chronological Retrospection
  & 64.55 & 55.45 & 55.45 & --- \\
  & (38.18--88.18) & (34.52--76.36) & (39.09--70.00) & \\

Invalid Time Detection
  & 20.91 & 19.09 & 41.82 & --- \\
  & (0.00--43.64) & (1.82--41.82) & (25.45--59.09) & \\

Temporal Adaptivity
  & 18.18 & 23.64 & 84.55 & --- \\
  & (2.73--36.36) & (9.09--40.91) & (76.36--91.82) & \\

Temporal Contextual Awareness
  & 32.73 & 42.73 & 78.18 & --- \\
  & (8.18--59.09) & (18.18--69.09) & (63.64--90.00) & \\

Temporal Flow Anomaly Detection
  & 0.00 & 1.82 & 66.36 & --- \\
  & (0.00--0.00) & (0.00--4.55) & (48.18--83.64) & \\

Time Gap Awareness
  & 0.00 & 4.55 & 61.82 & --- \\
  & (0.00--0.00) & (0.00--10.00) & (43.64--79.09) & \\

Timezone Sensitivity
  & 70.00 & 93.64 & 65.45 & --- \\
  & (49.09--87.27) & (90.00--97.27) & (50.91--80.00) & \\

\hline

Mean Output Tokens / Run
  & 563.03 & 1514.66 & 317.21 & $<$1e--08 \\
  & (486.58--641.05) & (1270.89--1785.34) & (283.39--353.33) & \\

Mean Thinking Tokens / Run
  & 0.00 & 887.42 & 91.19 & $<$1e--08 \\
  & (0.00--0.00) & (683.44--1127.52) & (76.67--111.07) & \\

Mean \# Think Blocks / Run
  & 0.00 & 0.99 & 1.90 & 1.3e--08 \\
  & (0.00--0.00) & (0.98--1.00) & (1.48--2.55) & \\

\% Runs w/ Think Blocks
  & 0.00 & 99.09 & 91.17 & --- \\
  & (0.00--0.00) & (98.05--99.87) & (86.36--95.45) & \\

Think Block Position in \% (S/M/E)
  & --- & 100.0 / 0.0 / 0.0 & 21.85 / 77.95 / 0.21 & --- \\

\hline

\% Heavy Markdown
  & 83.51 & 86.36 & 61.82 & --- \\
  & (75.84--90.26) & (79.48--92.34) & (52.60--70.78) & \\

\% Light Markdown
  & 95.58 & 97.40 & 86.75 & --- \\
  & (90.78--99.35) & (94.16--99.61) & (81.43--91.43) & \\

\hline

\% Any Degeneracy
  & 4.16 & 15.45 & 2.08 & 5.3e--05 \\
  & (1.56--7.66) & (9.09--22.21) & (0.65--3.77) & \\

\% Malformed Outputs
  & 0.00 & 0.91 & 0.26 & --- \\
  & (0.00--0.00) & (0.13--2.08) & (0.00--0.65) & \\

\% Infinite Repetitions
  & 4.16 & 1.30 & 0.00 & --- \\
  & (1.69--7.66) & (0.13--2.99) & (0.00--0.00) & \\

\% Reasoning Leakage
  & 0.00 & 1.30 & 0.78 & --- \\
  & (0.00--0.00) & (0.39--2.47) & (0.00--1.82) & \\

\% Formatting Leakage
  & 0.00 & 14.29 & 1.17 & --- \\
  & (0.00--0.00) & (8.18--21.04) & (0.13--2.47) & \\

\hline
\end{tabular}

\vspace{4pt}
\begin{minipage}{0.9\textwidth}
\footnotesize
\textit{Note.} WSR = Wilcoxon signed-rank test, computed between Qwen3 thinking mode and \textsc{Time} across 77 scenario-level comparisons. WSR $p$-values are shown for aggregate and structural metrics. \\
Think-position percentages are computed over the total set of observed \texttt{<think>} blocks across runs. They measure the proportion of \texttt{<think>} blocks observed at the start, in the middle, or at the end of the response. S/M/E = start / middle / end. Because these values are included only as descriptive summaries of placement behavior, confidence intervals are not reported.
\end{minipage}
\end{table*}

\clearpage

\begin{table*}[t]
\centering
\caption*{Appendix F.1.4: Structural and behavioral metrics (32B Models)}
\setlength{\tabcolsep}{4pt}
\renewcommand{\arraystretch}{1.0}
\begin{tabular}{lcccc}
\hline
\textbf{Statistic} &
\multicolumn{2}{c}{\textbf{Qwen3-32B}} &
\textbf{\textsc{Time}-32B} &
\textbf{WSR $p$} \\
& \textbf{No-Thinking} & \textbf{Thinking} & & \\
\hline

Benchmark Score
  & 31.82 & 37.40 & 64.81 & 5.0e--07 \\
  & (25.71--38.31) & (31.56--43.51) & (58.18--71.17) & \\

\hline

Chronological Retrospection
  & 61.82 & 60.00 & 63.64 & --- \\
  & (38.18--83.64) & (41.82--77.27) & (45.45--80.00) & \\

Invalid Time Detection
  & 11.82 & 31.82 & 52.73 & --- \\
  & (2.73--25.45) & (13.64--51.82) & (32.73--71.82) & \\

Temporal Adaptivity
  & 20.00 & 26.36 & 76.36 & --- \\
  & (1.82--44.55) & (7.27--47.27) & (64.55--87.27) & \\

Temporal Contextual Awareness
  & 40.00 & 43.64 & 76.36 & --- \\
  & (13.64--67.27) & (20.00--69.09) & (58.18--92.73) & \\

Temporal Flow Anomaly Detection
  & 0.91 & 3.64 & 44.55 & --- \\
  & (0.00--2.73) & (0.00--8.18) & (23.64--65.45) & \\

Time Gap Awareness
  & 0.91 & 3.64 & 58.18 & --- \\
  & (0.00--2.73) & (0.00--7.27) & (40.00--75.45) & \\

Timezone Sensitivity
  & 87.27 & 92.73 & 81.82 & --- \\
  & (78.18--94.55) & (83.64--99.09) & (68.18--92.73) & \\

\hline

Mean Output Tokens / Run
  & 608.96 & 1573.47 & 332.64 & $<$1e--08 \\
  & (533.37--686.50) & (1327.94--1856.57) & (296.15--371.77) & \\

Mean Thinking Tokens / Run
  & 0.00 & 910.52 & 84.16 & $<$1e--08 \\
  & (0.00--0.00) & (705.34--1158.80) & (71.18--98.27) & \\

Mean \# Think Blocks / Run
  & 0.00 & 0.99 & 1.67 & 1.4e--04 \\
  & (0.00--0.00) & (0.98--1.00) & (1.34--2.06) & \\

\% Runs w/ Think Blocks
  & 0.00 & 99.22 & 80.65 & --- \\
  & (0.00--0.00) & (97.92--100.00) & (73.51--87.40) & \\

Think Block Position in \% (S/M/E)
  & --- & 100.0 / 0.0 / 0.0 & 24.14 / 75.62 / 0.23 & --- \\

\hline

\% Heavy Markdown
  & 83.64 & 90.91 & 62.73 & --- \\
  & (77.01--89.61) & (85.19--95.71) & (53.51--71.69) & \\

\% Light Markdown
  & 95.84 & 98.31 & 85.45 & --- \\
  & (92.60--98.31) & (95.45--99.87) & (79.74--90.65) & \\

\hline

\% Any Degeneracy
  & 4.42 & 18.18 & 3.64 & 4.2e--04 \\
  & (2.60--6.49) & (11.43--25.32) & (1.95--5.97) & \\

\% Malformed Outputs
  & 0.00 & 0.78 & 1.95 & --- \\
  & (0.00--0.00) & (0.00--2.08) & (1.04--2.99) & \\

\% Infinite Repetitions
  & 4.29 & 0.91 & 1.30 & --- \\
  & (2.47--6.23) & (0.00--2.47) & (0.52--2.34) & \\

\% Reasoning Leakage
  & 0.13 & 0.91 & 1.17 & --- \\
  & (0.00--0.39) & (0.13--1.95) & (0.52--1.95) & \\

\% Formatting Leakage
  & 0.00 & 17.01 & 1.17 & --- \\
  & (0.00--0.00) & (10.39--24.16) & (0.13--2.86) & \\

\hline
\end{tabular}

\vspace{4pt}
\begin{minipage}{0.9\textwidth}
\footnotesize
\textit{Note.} WSR = Wilcoxon signed-rank test, computed between Qwen3 thinking mode and \textsc{Time} across 77 scenario-level comparisons. WSR $p$-values are shown for aggregate and structural metrics. \\
Think-position percentages are computed over the total set of observed \texttt{<think>} blocks across runs. They measure the proportion of \texttt{<think>} blocks observed at the start, in the middle, or at the end of the response. S/M/E = start / middle / end. Because these values are included only as descriptive summaries of placement behavior, confidence intervals are not reported.
\end{minipage}
\end{table*}

\clearpage

\begin{table*}[t]
\centering
\caption*{Appendix F.2: Phase-wise ablation details (32B Models)}
\setlength{\tabcolsep}{6pt}
\renewcommand{\arraystretch}{1.25}
\begin{tabularx}{\textwidth}{
  >{\raggedright\arraybackslash\hyphenpenalty=10000\exhyphenpenalty=10000}X
  ccccccc
}

\hline
\textbf{Statistic} &
\multicolumn{2}{c}{\textbf{Qwen3 32B}} &
\textbf{Phase 1} &
\textbf{Phase 2} &
\textbf{Phase 3} &
\textbf{\textsc{Time}} \\
&
\textbf{No-Thinking} &
\textbf{Thinking} &
&
&
&
\textbf{(32B)} \\
\hline

Benchmark Score
 & 31.82 & 37.40 & 42.47 & 56.88 & 52.08 & 64.81 \\

\hline

Chronological Retrospection
 & 61.82 & 60.00 & 63.64 & 52.73 & 65.45 & 63.64 \\

Invalid Time Detection
 & 11.82 & 31.82 & 39.09 & 21.82 & 30.91 & 52.73 \\

Temporal Adaptivity
 & 20.00 & 26.36 & 40.91 & 80.00 & 68.18 & 76.36 \\

Temporal Contextual Awareness
 & 40.00 & 43.64 & 48.18 & 50.00 & 49.09 & 76.36 \\

Temporal Flow Anomaly Detection
 & 0.91 & 3.64 & 7.27 & 49.09 & 40.00 & 44.55 \\

Time Gap Awareness
 & 0.91 & 3.64 & 3.64 & 59.09 & 44.55 & 58.18 \\

Timezone Sensitivity
 & 87.27 & 92.73 & 94.55 & 85.45 & 66.36 & 81.82 \\

\hline

Mean Output Tokens / Run
 & 608.96 & 1573.47 & 1434.56 & 362.45 & 294.51 & 332.64 \\

Mean Thinking Tokens / Run
 & 0.00 & 910.52 & 803.52 & 76.59 & 52.94 & 84.16 \\

Mean \# Think Blocks / Run
 & 0.00 & 0.99 & 0.99 & 1.12 & 1.25 & 1.67 \\

\% Runs w/ Think Blocks
 & 0.00 & 99.22 & 99.48 & 95.58 & 89.22 & 80.65 \\

 \% Think Blocks at Start
 & --- & 100.0 & 100.0 & 70.65 & 54.97 & 24.14 \\

\% Think Blocks in Middle
 & --- & 0.0 & 0.0 & 29.12 & 44.62 & 75.62 \\

\% Think Blocks at End
 & --- & 0.0 & 0.0 & 0.23 & 0.41 & 0.23 \\

\hline

\% Heavy Markdown
 & 83.64 & 90.91 & 88.57 & 65.32 & 61.43 & 62.73 \\

\% Light Markdown
 & 95.84 & 98.31 & 97.92 & 83.38 & 79.48 & 85.45 \\

\hline

\% Any Degeneracy
 & 4.42 & 18.18 & 13.90 & 4.68 & 0.78 & 3.64 \\

\% Malformed Outputs
 & 0.00 & 0.78 & 0.52 & 1.82 & 0.26 & 1.95 \\

\% Infinite Repetitions
 & 4.29 & 0.91 & 0.52 & 0.39 & 0.13 & 1.30 \\

\% Reasoning Leakage
 & 0.13 & 0.91 & 1.04 & 2.47 & 0.26 & 1.17 \\

\% Formatting Leakage
 & 0.00 & 17.01 & 12.47 & 1.17 & 0.26 & 1.17 \\

\hline
\end{tabularx}
\end{table*}

\end{document}